%% file: main_arxiv.tex
\documentclass[10pt,twocolumn,letterpaper]{article}

\usepackage{iccv}
\usepackage{times}
\usepackage{epsfig}
\usepackage{graphicx}
\usepackage{amsmath}
\usepackage{amssymb}
\usepackage{booktabs}
\usepackage{multirow}
\usepackage{multicol}
\usepackage{color}
\usepackage{xcolor}
\usepackage{mathtools}
\usepackage{subcaption}
\usepackage{makecell}
\usepackage{pifont}
\newcommand{\cmark}{\ding{51}}%
\newcommand{\xmark}{\ding{55}}%

\usepackage{mathtools}  
\usepackage{diffcoeff}  
\usepackage{color,colortbl}
\definecolor{Gray1}{gray}{0.95}
\definecolor{Gray2}{gray}{0.99}
\newcommand{\CC}[1]{\cellcolor{gray!#1}}
\newcommand\CCG[1][]{\CC{10}}
\newcommand{\seblock}{SE-Block\xspace}
\renewcommand{\cmark}{\color{blue}\ding{51}}%
\renewcommand{\xmark}{\color{red}\ding{55}}%

\DeclarePairedDelimiter\norm{\lVert}{\rVert}

\usepackage[pagebackref=true,breaklinks=true,letterpaper=true,colorlinks,bookmarks=false]{hyperref}

\iccvfinalcopy 

\def\iccvPaperID{1066} 

\ificcvfinal\fi

\begin{document}
\title{Mind the Backbone: Minimizing Backbone Distortion for\\ Robust Object Detection}

\author{Kuniaki Saito$^{1}$, Donghyun Kim$^{2}$, Piotr Teterwak$^{1}$, Rogerio Feris$^{2}$, Kate Saenko$^{1,3}$ \\
$^{1}$Boston University, $^{2}$MIT-IBM Watson AI Lab , $^{3}$Meta AI\\
\tt\small \{keisaito, piotrt, saenko\}@bu.edu, dkim@ibm.com, rsferis@us.ibm.com
}

\maketitle
\ificcvfinal\thispagestyle{empty}\fi

\input{paragraphs/abstract.tex}

\input{paragraphs/intro.tex}

\input{paragraphs/related.tex}
\input{paragraphs/method.tex}

\input{paragraphs/experiments.tex}

\input{paragraphs/conclusion.tex}


\setcounter{section}{0}
\setcounter{table}{0}
\setcounter{figure}{0}

\def\thesection{\Alph{section}}
\renewcommand{\thetable}{\Alph{table}}
\renewcommand{\thefigure}{\Alph{figure}}
\appendix

\input{paragraphs/supple.tex}

{\small
\bibliographystyle{ieee_fullname}
\bibliography{egbib}
}

\end{document}

%% file: paragraphs/abstract.tex

\begin{abstract}
Building object detectors that are robust to domain shifts is critical for real-world applications. Prior approaches fine-tune a pre-trained backbone and risk overfitting it to in-distribution (ID) data and distorting  features useful for out-of-distribution (OOD) generalization. We propose to use Relative Gradient Norm (RGN) as a way to measure the vulnerability of a backbone to feature distortion, and show that high RGN is indeed correlated with lower OOD performance. Our analysis of RGN yields interesting findings: some backbones lose OOD robustness during fine-tuning, but others gain robustness because their architecture prevents the parameters from changing too much from the initial model. 
Given these findings, we present recipes to boost OOD robustness for both types of backbones. Specifically, we investigate regularization and architectural choices for minimizing gradient updates so as to prevent the tuned backbone from losing generalizable features. Our proposed techniques complement each other and show substantial improvements over baselines on diverse architectures and datasets.
\end{abstract}

%% file: paragraphs/intro.tex
\section{Introduction}
Fine-tuning a large-scale pre-trained vision model has become the defacto transfer learning paradigm that significantly improves performance on diverse downstream tasks compared to training from scratch~\cite{girshick2014rich,long2015fully,chu2016best,yosinski2014transferable}. Although naive fine-tuning performs well on the downstream task's training distribution (in-distribution; ID), it has subpar performance on data unseen during training (out-of-distribution; OOD)~\cite{kumar2022fine}. 
Kumar \etal~\cite{kumar2022fine} suggest that fine-tuning without unnecessarily distorting the pre-trained backbone is the key to achieving high performance in both ID and OOD. They show that ``linear warm-up'' training, \ie training a linear head first while freezing the backbone before fine-tuning all layers, is effective for robust image classification. 
\input{figures/teaser.tex}
Inspired by this, we explore the effect of backbone distortion during fine-tuning in object detection. Although there exist generalizable detection approaches that use data augmentation~\cite{chen2021robust}, feature augmentation~\cite{fan2022normalization}, or advanced objectives~\cite{wang2021robust}, our work is the first to ask whether biasing toward the initial model is effective for this task.

We study an equivalent of ``linear warm-up'' training for object detection. Specifically, we first train the decoder (all detector modules except for the backbone) while keeping the backbone frozen, which we refer to as \textit{Decoder-Probing (DP)}. After DP, we perform full fine-tuning of the detector (including the backbone), which we call \textit{Decoder-Probing Fine-Tuning (DP-FT)}. Surprisingly, we observe that the improvement in OOD performance of DP-FT over DP depends on the architecture of the backbone.
To analyze this phenomenon, we investigate the relationship between performance improvement from fine-tuning and the amount of gradient in each parameter. Specifically, two detectors are compared for each backbone to compute the performance change after tuning the backbone: (1) a DP model and (2) fine-tuning all modules (FT). We then compute the ratio of gradient norm to parameter norm, i.e., \textit{Relative Gradient Norm (RGN)}~\cite{lee2022surgical}, calculated on the training (in-distribution) data using DP. Intuitively, RGN indicates how much updating feature extractors require to fit ID data. We measure RGN's correlation with the performance improvement on OOD relative to that on ID.  
Our novel findings are three-fold: (1) a model with small RGN is more likely to improve performance on both OOD and ID data after fine-tuning, (2)  RGN depends on the network architecture, (3) in particular, some modules such as squeeze excitation (SE) blocks~\cite{hu2018squeeze} can reduce the gradient, thus lowering RGN and improving both OOD and ID performance after fine-tuning. 

Based on these findings, we explore two techniques to preserve the generalizability of a diverse set of pre-trained models (see Fig.~\ref{fig:teaser}).
The first is a regularizer that minimizes the distance from the initial model in the parameter space to prevent feature distortion. The backbone is constrained to find a point performing well on ID, yet not too far away from the initial model. This simple regularization effectively boosts OOD performance for diverse architectures, yet is overlooked in generalizable object detection. Second, we study the effect of the decoder's architecture and training iterations on decreasing the RGN of the backbone. A decoder producing a lower RGN tends to train a detector more generalizable to OOD. Specifically, inserting an \seblock into the backbone significantly decreases RGN. 

Our contributions can be summarized as follows:
\vspace{-2mm}
\begin{itemize}
    \item We reveal a novel finding that some backbones lose more OOD robustness after fine-tuning than others. At the same time, some backbones can significantly increase robustness due to an architecture that prevents the parameters from being too far away from the initial model, such as having an \seblock.
    \vspace{-1mm}
    \item We present methods to boost OOD robustness for both types of backbones. Specifically,  weight regularization and  decoder design are investigated, and the two techniques are shown to complement each other.  
    \vspace{-1mm}
    \item Our methods show substantial improvements  over baselines when evaluating on diverse architectures and object detection datasets. 
\end{itemize}

%% file: figures/teaser.tex
\begin{figure}
    \centering
    \includegraphics[width=\linewidth]{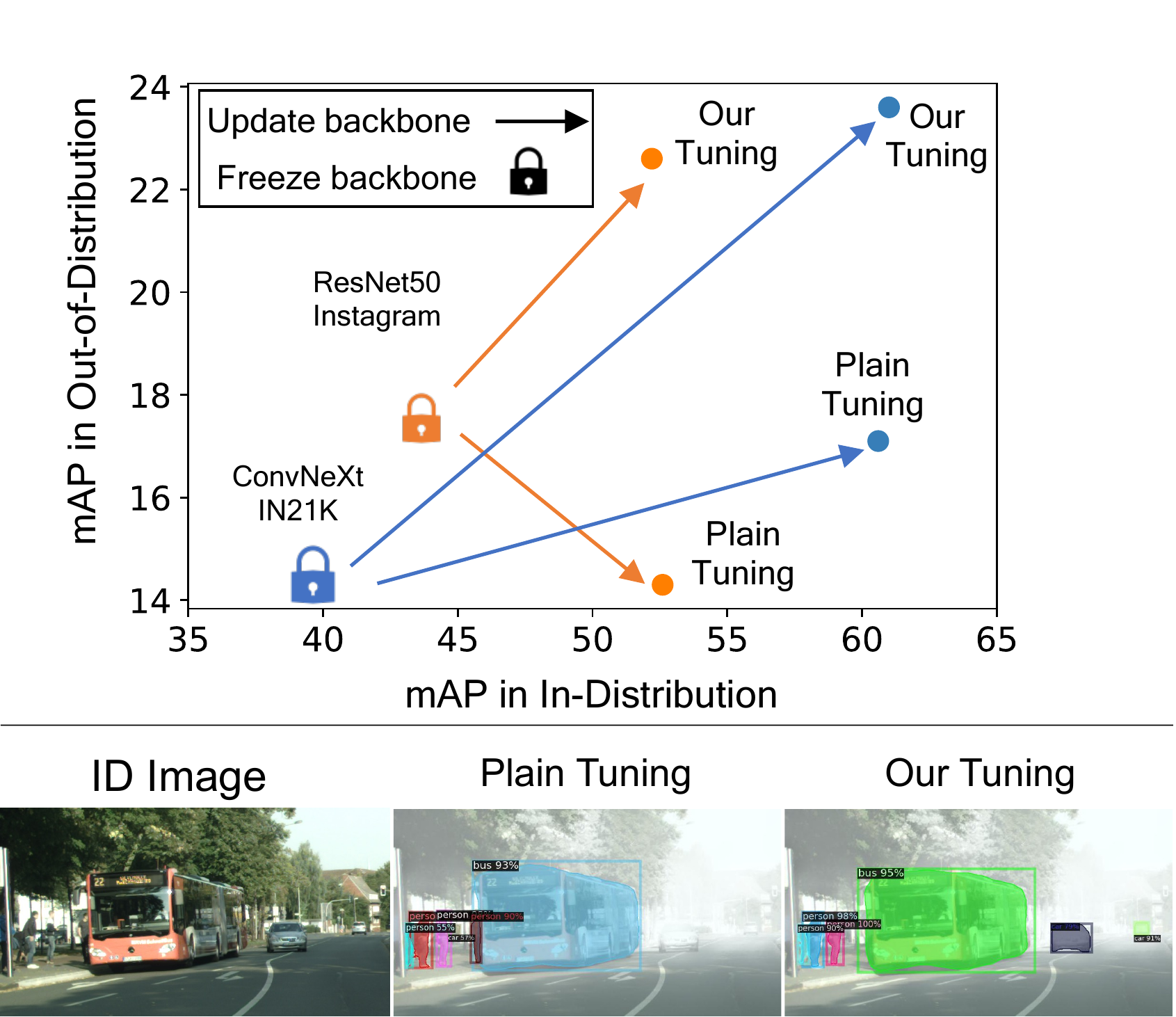}
    \vspace{-4mm}
    \caption{Top: The standard fine-tuning approach in object detection can cause overfitting to ID and  hurt the performance on OOD data (\textcolor{orange}{Orange} dot with \textbf{Plain Tuning}) or achieve minimal improvements (\textcolor{blue}{Blue} dot with \textbf{Plain Tuning}) depending on the network architecture. Based on a thorough analysis, we propose a new fine-tuning recipe that can achieve good performance on both ID and OOD performance for diverse pre-trained backbones (\textcolor{blue}{Blue} and \textcolor{orange}{Orange} dots with \textbf{Our Tuning}). Results are on Pascal~\cite{everingham2010pascal}, see Table~\ref{tb:seblock}. Bottom: We visualize detected results in Foggy Cityscape.
    } 
    \label{fig:teaser}
\end{figure}

%% file: paragraphs/related.tex
\section{Related Work}
\vspace{-2mm}
\noindent \textbf{Domain Generalization in Image classification.}
Robust fine-tuning, namely, adapting a pre-trained model to a downstream task without losing generalization to unseen domains, has been popular in recent work. Gulrajani \etal \cite{gulrajani2020search} find that simple ERM, \ie fine-tuning on source with well-tuned hyperparameters, shows strong generalization. Kim \etal~\cite{kim2022broad} and Angarano \etal~\cite{angarano2022back} show that larger models have stronger OOD performance. Kumar \etal propose warm-up with a linear head~\cite{kumar2022fine}, while Wortsman \etal propose to interpolate the weights of a fine-tuned and a zero-shot model to produce a single robust model~\cite{wortsman2022robust}. This technique is extended to ensemble diverse models trained from the same initialization~\cite{wortsman2022model,rame2022diverse}, and to ensemble several checkpoints from one training trajectory~\cite{cha2021swad}. Regularizing the change w.r.t the pre-trained model's features is another promising way to tackle the problem~\cite{chen2021contrastive, cha2022domain}. 
Surgical fine-tuning~\cite{lee2022surgical} changes only a subset of layers to adapt a model to a downstream task. Their goal is to improve the performance on the downstream task given a small number of training data. They find that tuning a block with a large RGN achieves high performance on the downstream task. In contrast, our findings suggest that large RGN can indicate the risk of overfitting the training data. 

\input{tables/dp-ft_compare.tex}

\noindent \textbf{Data Augmentation.}
Data augmentation effectively expands the training domain~\cite{zhou2021domain,xu2021fourier} \eg, Augmix~\cite{hendrycks2019augmix} exploits diverse synthetic data augmentation, and augmentation with style transfer can further diversify training images~\cite{zhong2022adversarial}. Our experiments show that the backbone can lose generalization even with augmentation, needing regularization to maintain OOD robustness. 

\noindent \textbf{Domain Generalization in Object detection.}
Wang~\etal employ temporal frames to ensure the consistency of features~\cite{wang2021robust}. Normalization perturbation (NP)~\cite{fan2022normalization}, concurrent with our work, uses style randomization in feature space. Det-AdvProp~\cite{chen2021robust} combines adversarial and clean examples to train a model. The settings of NP and Det-AdvProp are comparable to ours, but we do not compare against their approaches since they do not publish code to train models. Instead, we show the compatibility of our approach with data augmentation. This indicates the potential compatibility with their methods, which use data augmentation for training. 
There exist multiple source generalizable object detection~\cite{lin2021domain, wang2019towards}, but we tackle a single-source scenario with no domain labels. Vasconcelos~\etal~\cite{vasconcelos2022proper} demonstrate that training only a strong decoder on top of a frozen backbone can improve detection models. They focus on evaluating the model on ID, but we inspect the effectiveness of OOD generalization and fine-tuning the backbone. 

\noindent \textbf{Continual Learning.}
Our study connects with early work on continual learning~\cite{kirkpatrick2016overcoming} in considering the distance from a pre-trained model in weight space. They find that elastic weight consolidation (EWC), which accounts for the importance of each weight on the older task, is useful for maintaining performance on older tasks. Mirzadeh \etal~\cite{mirzadeh2022wide} show that wider neural networks forget less and have sparser gradient updates. 
Xuhong~\etal~\cite{xuhong2018explicit} inspect fine-tuning for retaining the features learned on the source task, compare several regularization methods including EWC, and observe that a simple L2 penalty on the pre-trained model mitigates overfitting well, which we also observe in experiments (see appendix for more details).
Wortsman~\etal~\cite{wortsman2022robust} also use the L2 penalty as a baseline in an image classification task but the penalty is not explored in object detection. We highlight that we are the first to show that weight regularization is an overlooked, yet powerful tool for keeping OOD performance in object detection. 

%% file: tables/dp-ft_compare.tex
\begin{table*}
\centering

\scalebox{0.98}{
\begin{tabular}{l|l|l|l|l|l}
\toprule

\multicolumn{1}{c|}{Backbone} & Tuning  & \multicolumn{1}{c|}{Comic} &  \multicolumn{1}{c|}{Water} &  \multicolumn{1}{c|}{Clipart} & \multicolumn{1}{c}{ID: Pascal}   \\\hline
\multirow{3}{*}{ResNet50 Instagram~\cite{mahajan2018exploring}} & DP    & 15.7  & 21.2       & 15.3    & 44.6 \\
& FT    & 7.5 (\textcolor{red}{-8.2})  & 19.4  (\textcolor{red}{-1.8})  & 11.4 (\textcolor{red}{-3.9})& 50.4  (\textcolor{blue}{+5.8}) \\
& DP-FT & 9.1 (\textcolor{red}{-6.6})   & 21.0   (\textcolor{red}{-0.2}) & 12.9 (\textcolor{red}{-2.4})  & 52.6  (\textcolor{blue}{+8.0}) \\ \hline
\multirow{3}{*}{EfficientNet-B2 JFT~\cite{xie2020self}}& DP    & 12.6  & 20.4       & 15.1    & 40.2 \\
& FT    & 17.1 (\textcolor{blue}{+4.5}) & 27.2  (\textcolor{blue}{+6.8})       & 18.0   (\textcolor{blue}{+2.9})   & 53.4 (\textcolor{blue}{+13.2})\\
        & DP-FT & 17.4  (\textcolor{blue}{+4.7})) & 29.4      (\textcolor{blue}{+9.0})   & 20.7  (\textcolor{blue}{+5.6})    & 55.3  (\textcolor{blue}{+15.1})\\\hline
\multirow{3}{*}{ConvNeXt Base IN21K~\cite{liu2022convnet}}                & DP    & 11.7  & 17.3       & 14.0    & 39.7 \\
& FT    & 11.5  (\textcolor{red}{-0.2})  & 22.9   (\textcolor{blue}{+5.6})  & 16.8  (\textcolor{blue}{+2.8})    & 60.6  (\textcolor{blue}{+20.9}) \\
                & DP-FT & 13.6  (\textcolor{blue}{+2.9})  & 24.7    (\textcolor{blue}{+7.4})     & 19.1   (\textcolor{blue}{+5.1})   & 62.3 (\textcolor{blue}{+22.6}) \\
\bottomrule
\end{tabular}
}
\vspace{-3mm}
\caption{\small Analysis of DP (Detector Probing, \ie tuning only the decoder, freezing backbone), FT (Fine-Tuning all modules), and DP-FT (tuning all layers after DP training) on the three OOD domains and ID (training) domain. Interestingly, EfficientNet-B2 FT improves OOD performance over DP while ResNet50 FT degrades OOD performance. Overall, DP-FT improves performance over FT. }
\label{tb:dp-ft}
\end{table*}

%% file: paragraphs/method.tex
\input{figures/per_layer_rgn.tex}
\vspace{-3mm}
\section{Analysis of Feature Distortion}~\label{sec:analysis}
In this section, we aim to answer three questions; (1) how the model's distortion can be quantified, (2) the correlation between OOD performance and the distortion, and (3) whether the distortion depends on the backbone architecture. 
We start by analyzing three tuning strategies; (1) a decoder-probing (DP) model, which freezes the feature extractor while training the remaining detector modules, \ie, decoder, (2) fine-tuning (FT) model, which tunes both the extractor and detector, (3) a decoder-probing before fine-tuning (DP-FT), which applies DP training before FT. 

\noindent \textbf{Set-up.} 
We use the Faster R-CNN model~\cite{faster} with a feature pyramid network~\cite{lin2017feature} as our detector since it is the typical detector architecture. We regard the pyramid modules and detector head as the decoder, wherein only the decoder is trained for a DP model. We follow the protocol of detectron2~\cite{wu2019detectron2} for hyper-parameter selection, \eg, the shallow CNN layers (stages 1 and 2) and all BatchNorm parameters are frozen by the default setting of the library. DP and FT train for the same number of iterations, DP-FT initializes the decoder with the DP model and trains both decoder and backbone as in FT training. 
Backbones are chosen mainly from models pre-trained with large-scale data such as Instagram~\cite{mahajan2018exploring}, JFT~\cite{sun2017revisiting}, and ImageNet21K~\cite{deng2009imagenet} (IN21K) since our primary goal is the investigation of robust models.
We utilize Pascal~\cite{everingham2010pascal} (16,551 training images) or Cityscapes~\cite{cordts2016cityscapes} (2,965 training images) as training data. We employ watercolor, comic, and clipart~\cite{inoue2018cross} as OOD in Pascal, FoggyCityscapes~\cite{sakaridis2018semantic} and BDD~\cite{yu2020bdd100k} as OOD in Cityscapes. 
We focus on the setting where the training data is not very large as this is the case in many applications. 
We employ the mean Average Precision (mAP) used in COCO~\cite{coco} as the evaluation metric.
\input{figures/rgn_vs_imp_tmp}

\noindent \textbf{The performance improvement by DP-FT depends on the architecture.}
First, we conduct experiments to see the effect of DP-FT in Table~\ref{tb:dp-ft}. 
Interestingly, the performance improvement over DP depends on the pre-trained models. EfficientNet~\cite{tan2019efficientnet,xie2020self} improves both OOD and ID performance by fine-tuning feature extractor, but ResNet50~\cite{mahajan2018exploring} significantly degrades OOD performance. Also, DP-FT improves performance compared to FT but still underperforms DP in OOD for ResNet50. Overall, ConvNeXt improves OOD performance, but the gain is marginal, considering its high performance on ID. Prior work on image classification~\cite{kumar2022fine} only reports that naive fine-tuning tends to decrease OOD generalization. Our novel finding is that the performance decrease or increase depends on the architecture of the pre-trained models. 

\noindent \textbf{Relative Gradient Norm differs by architectures.}
The analysis above motivates us to study the factors that improve OOD performance with simple fine-tuning. 
From the insight on linear-warm-up training for image classification~\cite{kumar2022fine}, if the amount of updates, \ie, model distortion, by fine-tuning is large, the model will likely lose generalization on OOD. To quantify the model distortion, we compute the variant of the ratio of gradient norm to parameter norm (RGN)~\cite{lee2022surgical}. 

Let $\mathbf{W} \in \mathbb{R}^{C_{i} \times C_{o} \times F} $ denote a weight parameter in a convolution layer, where $C_{i}, C_{o}, F$ denote input, output channel size, and the kernel size of filters. 
Since we are interested in the scale of updates per each filter, we take summation within each filter to compute RGN in layer $L$, $\mathbf{RGN}^{L}_{ij} = \frac{\sum_{k} |\nabla{\mathbf{W}_{ijk}}|}{\sum_{k} |\mathbf{W}_{ijk}|}$. Then, $\mathbf{RGN}^{L} = \frac{1}{C_{i}C_{o}}\sum_{i, j} \mathbf{RGN}^{L}_{ij}$. To get the RGN value for one model, we take the average over layers, $L$.
We utilize the DP model to compute the gradient of detection loss; thus, RGN indicates how aggressively $W$ will be updated by fine-tuning. RGN is averaged over training images. Note that since we do not update the DP model during this computation. RGN shows the expected update at the starting point of DP-FT. 

Fig.~\ref{fig:rgn_depth} plots RGN values for the three models on Pascal, with different layers, used in Table~\ref{tb:dp-ft}. RGN differs significantly by model; EfficientNet has a small RGN in most layers, ConvNeXt has a large RGN in the first few layers, and ResNet50 has large RGN  across all layers. Also, note that ResNet50 loses OOD generalization by fine-tuning while the other two networks improve it in Table~\ref{tb:dp-ft}. The RGN averaged over all layers are 1.34 (ResNet50), 0.13 (ConvNeXt), and 0.06 (EfficientNet), respectively.
This observation is consistent with previous findings, \ie, the performance on OOD will be high if a pre-trained model does not have to aggressively update parameters to fit a downstream task~\cite{kumar2022fine}. 
However, our novel finding is that the model distortion is specific to each pre-trained model. Next, we conduct a more extensive study to investigate whether the difference stems from architecture or the dataset used to train the model.

\noindent \textbf{Relative Gradient Norm (RGN) has a negative correlation with performance improvement in OOD.}
In Fig.~\ref{fig:rgn_vs_improvement}, we investigate the relationship between RGN and the ratio of improvement on OOD to improvement on ID, \ie $\frac{(\rm{OOD_{FT}}-OOD\rm{_{DP}})}{(\rm{ID_{FT}}-ID_{DP})}$, where $\rm{OOD}$ and $\rm{ID}$ denote the mAP on OOD and ID respectively. We employ 14 models with diverse architectures and different pre-training datasets, \eg, ResNet~\cite{he2016deep}, SeNet~\cite{hu2018squeeze}, EfficientNet~\cite{tan2019efficientnet}, ConvNeXt~\cite{liu2022convnet}, MobileNetV2~\cite{sandler2018mobilenetv2}.
We observe that RGN and relative improvement has a negative correlation, indicating that RGN is a good measurement of the performance improvement on OOD gained by fine-tuning on ID. Intuitively, the target domain with low mAP is far from the ID. Then, the correlation becomes more evident in these plots as OOD gets farther away from ID. For example, the correlation is more evident in Comic (mAP: 10.3) while it is not evident in Watercolor (map: 22.0). 

\input{tables/rgn_dataset_architecture.tex}
\noindent \textbf{Pre-training dataset and SE-Block affect RGN.}
In Table~\ref{tb:rgn_dataset_arch}, we pick ResNet50 trained with different datasets and SeNet50 to show their RGN. First, surprisingly, using more data tends to increase RGN, as shown in the results of ResNet50. If a model is trained on diverse data, it may need to lose much representation to be specific to the downstream task. Second, SeNet has a much smaller RGN than ResNet50 does. The only difference between SeNet and ResNet50 is the existence of \seblock. Therefore, we conjecture that one of the keys to reducing RGN is in \seblock and investigate it in the next paragraph. 

\input{figures/seblock_histogram.tex}
\noindent \textbf{Analysis of \seblock.}
Squeeze-and-excitation block performs channel-wise attention in the residual block as follows:
\begin{equation}
     \hat{\mathbf{X}} = \mathbf{X} + \sigma(S(R(\mathbf{X}))) \circ R(\mathbf{X}), 
\end{equation}
where $\sigma$, $S$, $R$, and $\circ$ denote sigmoid activation, multi-layer perceptron after average pooling, convolution layers, and channel-wise attention, respectively. This block applies sigmoid activation to scale outputs from 0 to 1. Therefore, the block masks several outputs from $R(X)$ resulting in a sparse activation pattern. In Fig.~\ref{fig:se_histogram}, we visualize the histogram of the activation from four SE-Blocks, computed on the SeNet~\cite{hu2018squeeze} pre-trained on ImageNet~\cite{deng2009imagenet}. We observe that many units are close to either zero or one.

For simplicity, let $M(\mathbf{X})$ denote $S(R(\mathbf{X}))$. Then, we can calculate the derivative as follows:
\begin{equation}\label{eq:se_derivative}
    \diffp{\hat{\mathbf{X}}}{\mathbf{X}}= I + \diffp{\sigma(M(\mathbf{X}))}{\mathbf{X}} \circ R(\mathbf{X}) + \sigma(M(\mathbf{X}))\circ \diffp{R(\mathbf{X})}{\mathbf{X}}.
\end{equation}

Note that the second term includes a derivative of the sigmoid activation, which is nearly zero if the activation is close to one or zero:
\begin{equation}
   \diffp{\sigma(M(\mathbf{X}))}{\mathbf{X}}= (I - \sigma(M(\mathbf{X})))\sigma(M(\mathbf{X})) \diffp{M(\mathbf{X})}{\mathbf{X}}.
\end{equation}
Also, $\sigma(M(\mathbf{X})))$ in the third term of Eq. ~\ref{eq:se_derivative} can mask the derivative from $R(\mathbf{X})$. From this analysis and empirical findings, we hypothesize that \seblock is masking gradient from upper layers and promotes robust fine-tuning. 

In summary, in this analysis, we introduce RGN to measure the distortion of the model by fine-tuning, revealing that the large distortion is correlated with lower OOD performance.  Also, some backbones lose OOD robustness during fine-tuning, but others gain robustness because their architecture prevents distortion. 

\section{Regularization for Robust Fine-tuning}
Given the findings in the previous section, we aim to find a fine-tuning recipe for achieving a robust object detector.
In the previous section, we have seen that large RGN distorts features of the pre-trained model, and \seblock can effectively reduce the amount of gradient. In addition, DP-FT boosts OOD performance compared to FT. 
We build two techniques inspired by these observations on top of DP-FT. 
\input{figures/training_figure.tex}

\input{tables/main_results.tex}

\subsection{Weight Regularization}
Our analysis shows that RGN is a key factor in losing generalizability. To reduce RGN and incorporate an explicit inductive bias of a pre-trained model, we investigate the effect of introducing the distance from the initial value as a regularizer, which we denote $\Omega(w)$, and optimize it with detection loss as follows:

\begin{equation}\label{eq:objective}
    L(w) = L_{det}(w) + \lambda\Omega(w),
\end{equation}
where $L_{det}(w)$ denotes the loss of object detection. Note that the decoder modules are trained without regularization. 
There are several options for $\Omega(w)$ as indicated by previous work on transfer learning~\cite{kirkpatrick2016overcoming,xuhong2018explicit}. 

\noindent \textbf{$L^{2}$ penalty.}
One simple option for the distance function is the L2 distance,
\begin{equation}
\Omega(w) = \sum_{i}\norm{w_{i}-{w_{i}}^{pre}}^{2}.
\end{equation}
The distance looks naive, yet it empirically works well to preserve robustness to OOD. 

\noindent \textbf{$L^{2}$ penalty weighted by RGN.}
We also study weighting the $L^{2}$ penalty by RGN:
\begin{equation}
\Omega(w) =  \sum_{i}{{\mathbf{RGN}}_{i}} \norm{w_{i}-{w_{i}}^{pre}}^{2}, 
\end{equation}
where $\mathbf{RGN}_{i}$ denotes RGN value for the parameter $w_{i}$. The RGN value needs to be computed after DP training. 

\noindent \textbf{Elastic weight consolidation (EWC).}
EWC~\cite{kirkpatrick2016overcoming} utilizes a fisher-information matrix to weight the regularizer, 
\begin{equation}
\Omega(w) =  \sum_{i}{{F}_{i}} \norm{w_{i}-{w_{i}}^{pre}}^{2}, 
\end{equation}
where $F$ denotes the fisher matrix. Since Kirkpatrick~\etal~\cite{kirkpatrick2016overcoming} employ diagonal approximation to compute the Fisher matrix, the biggest difference between the penalty with RGN and this penalty is in the normalization by the norm of the parameter. 
Empirically, we find that these techniques are almost equally effective in boosting the performance of OOD, that is, by choosing proper $\lambda$, they show comparable performance in both OOD and ID. 
But, the regularization with RGN excels at transferring $\lambda$ across different pre-trained models probably because RGN considers the scale of gradient and weights, making regularization's strength consistent across different architectures. In experiments, we employ the regularization with RGN and leave the comparison among regularizers in the appendix.

\subsection{Regularization by Decoder Design}\label{sec:decoder}
\noindent\textbf{Decoder-Probing with \seblock.}
Although the \seblock effectively avoids distorting features, not all pre-trained models are built with it. Thus, we study inserting  SE-Blocks into several layers of pre-trained models and perform decoder-probing fine-tuning. By default, ResNet-like architectures split the network into four stages. We insert the block at the end of each stage, as shown in Fig.~\ref{fig:seblock_figure}. We tune the decoder and \seblock during decoder-probing training, then tune all modules during fine-tuning. The block is expected to filter the gradient from the upper layers, thus leading to better OOD performance. Additionally, since the block is a lightweight plug-in module, the parameter and running time increase during inference is very small. 

\noindent\textbf{Stronger decoder.}
One simple way to reduce the gradient updates to the backbone model is to use a strong decoder in decoder-probing training, then fine-tune the whole network. If the strong decoder decreases the loss, the backbone will not be significantly updated. In experiments, we investigate increasing training iterations in decoder-probing and decoder architecture. 

%% file: figures/per_layer_rgn.tex
\begin{figure}
    \centering
    \includegraphics[width=0.7\linewidth]{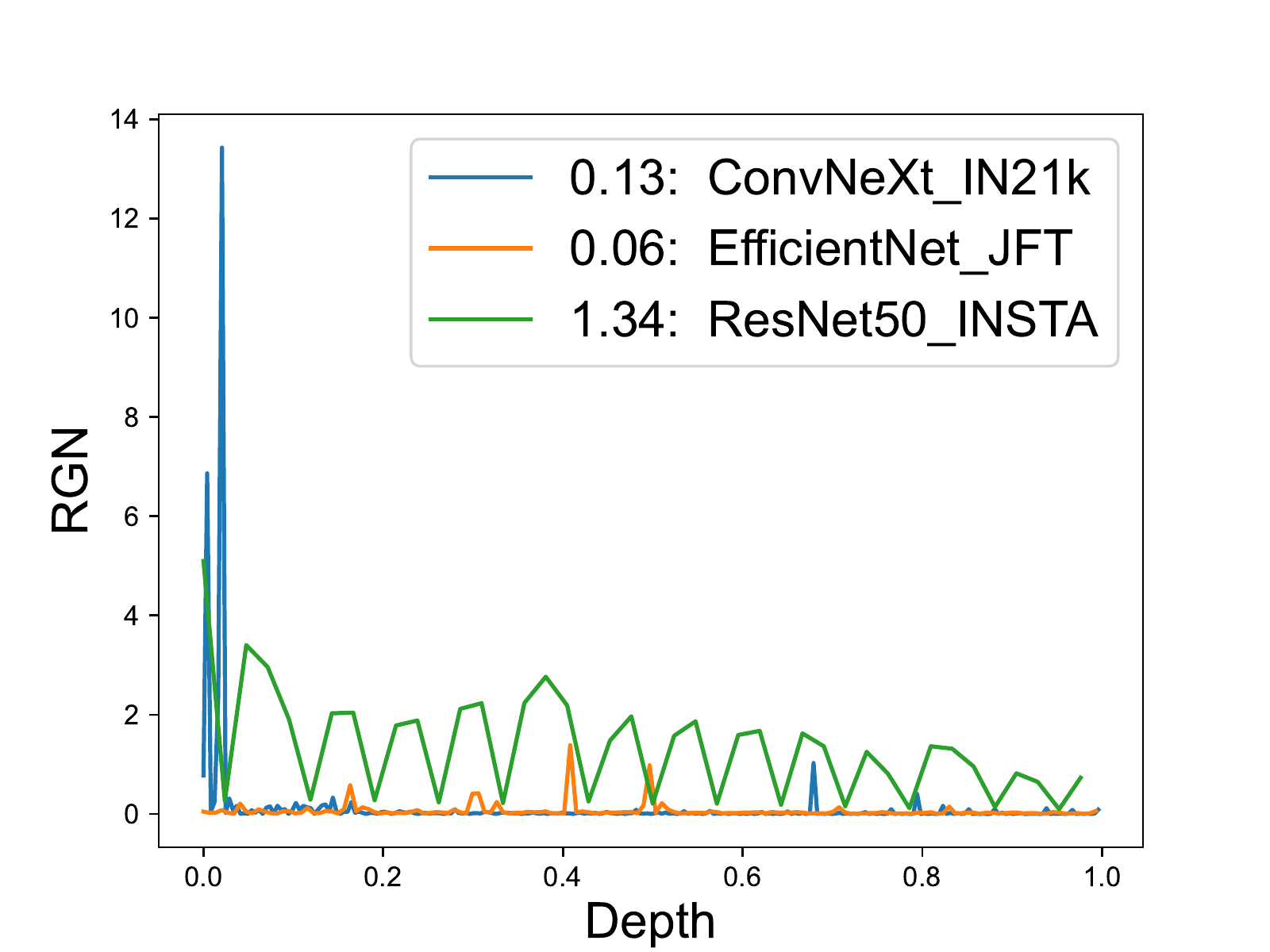}
    \vspace{-3mm}
    \caption{\textbf{RGN of three DP models on Pascal.} The depth is normalized by the total number of layers. Interestingly, RGN values differ a lot by models (legend shows RGN averaged over layers). ResNet50 has a large RGN in all layers, while EfficientNet has a very small RGN in all layers, which seems to support the generalization in OOD.}
    \label{fig:rgn_depth}
\end{figure}

%% file: figures/rgn_vs_imp_tmp.tex
\begin{figure*}[t]
 \centering
\begin{subfigure}{.3\textwidth}
  \centering
  \includegraphics[width=\linewidth]{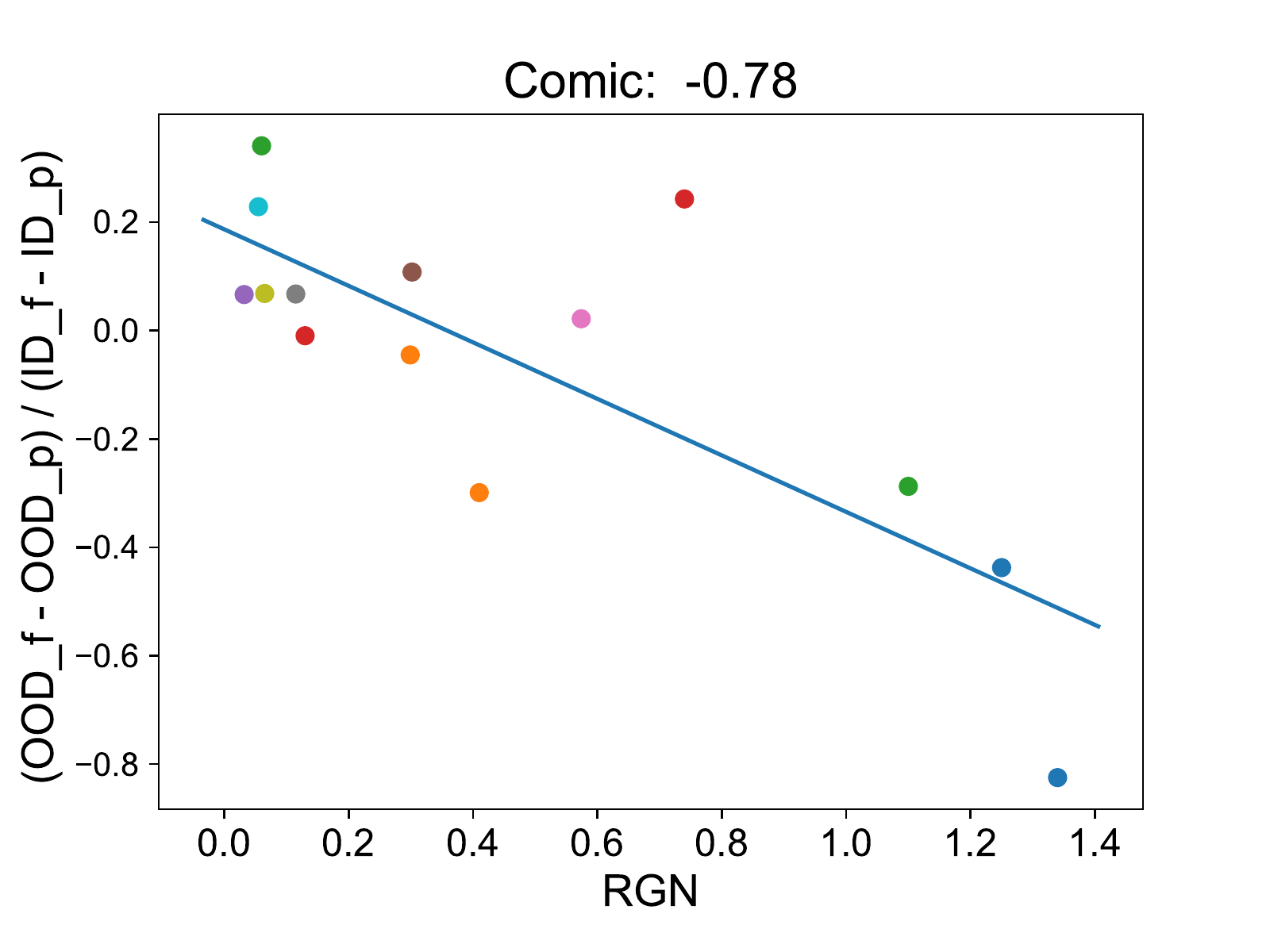}
  \caption{Comic: -0.78 (mAP 10.3)}
  \label{fig:comic}
\end{subfigure}%
\begin{subfigure}{0.3\textwidth}
  \centering
  \includegraphics[width=\linewidth]{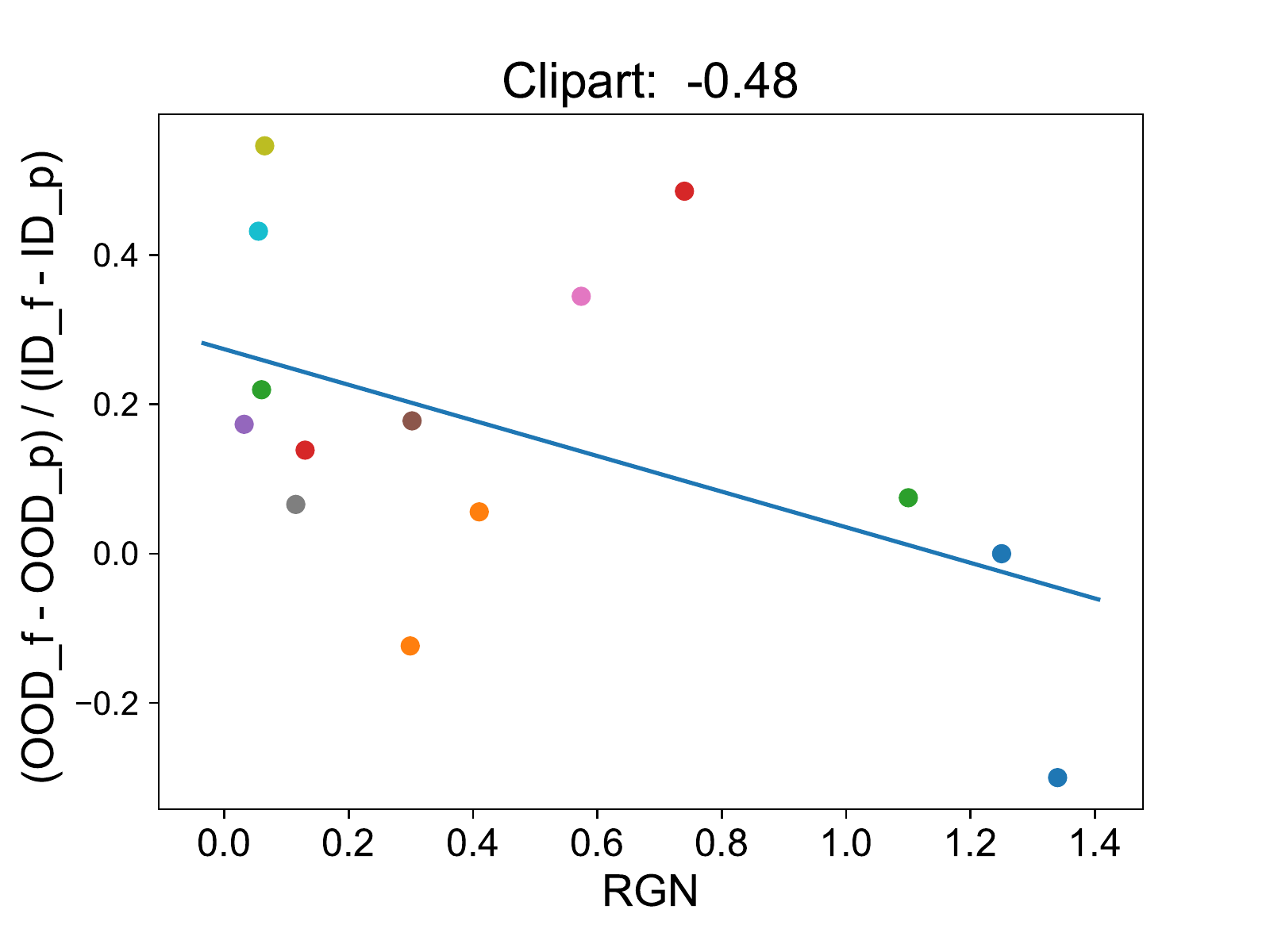}
  \caption{Clipart: -0.48 (mAP 14.8)}
  \label{fig:clipart}
\end{subfigure}
\begin{subfigure}{0.3\textwidth}
  \centering
  \includegraphics[width=\linewidth]{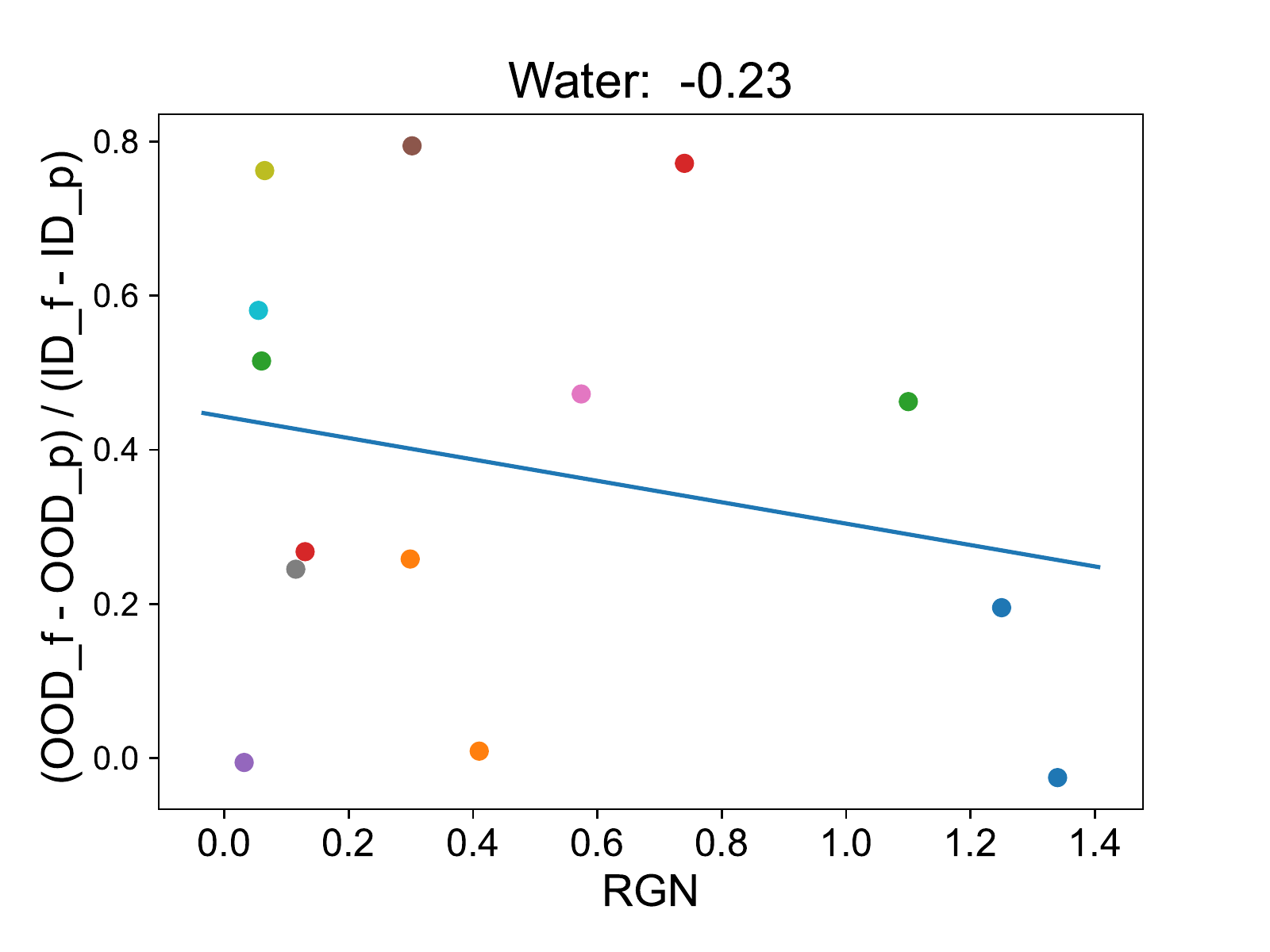}
  \caption{Watercolor: -0.23 (mAP 22.0)}
  \label{fig:water}
\end{subfigure}\par 
\begin{subfigure}{0.3\textwidth}
  \centering
  \includegraphics[width=\linewidth]{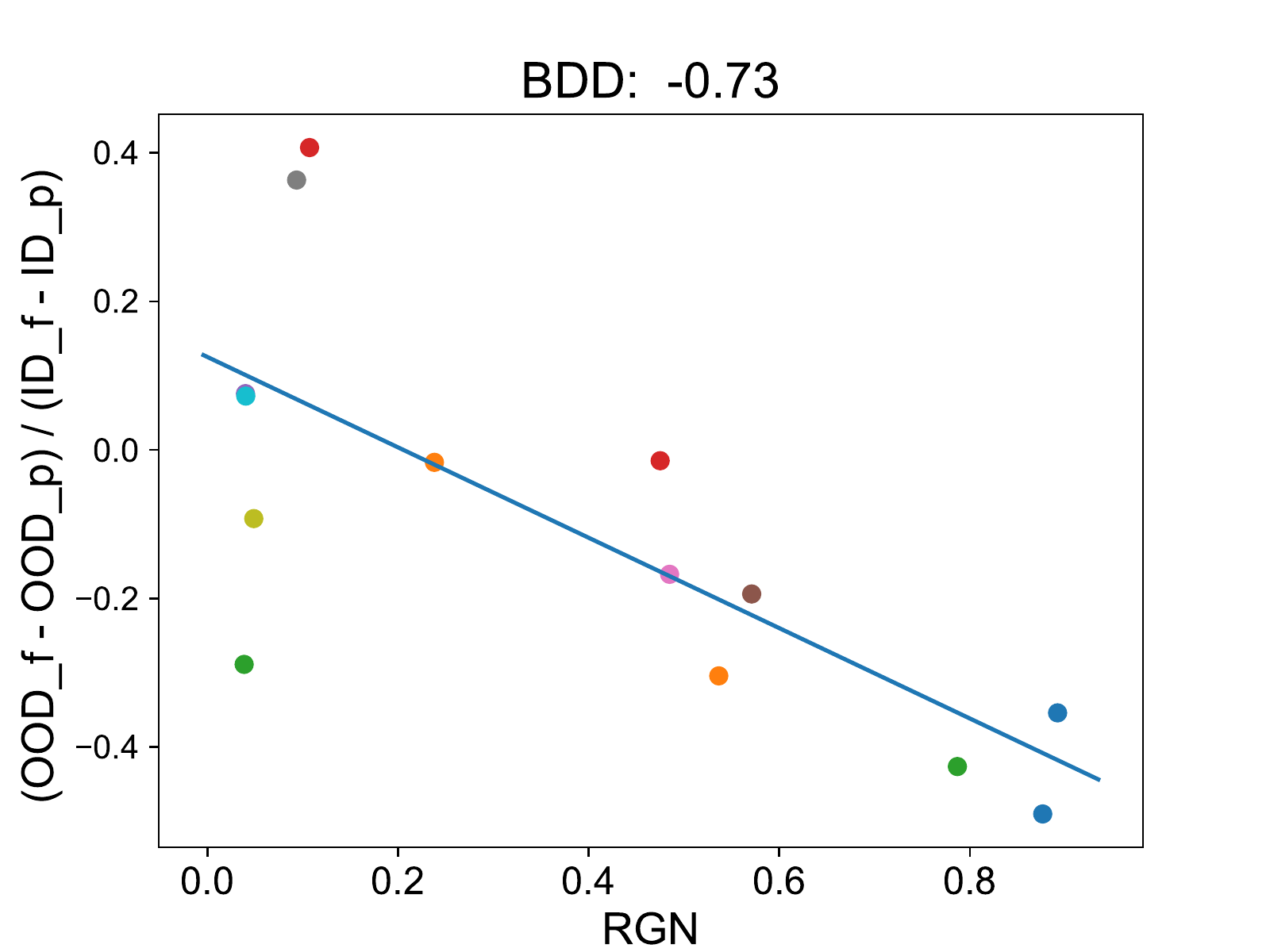}
  \caption{BDD: -0.73 (mAP 5.9)}
  \label{fig:bdd}
\end{subfigure}
\begin{subfigure}{0.3\textwidth}
  \centering
  \includegraphics[width=\linewidth]{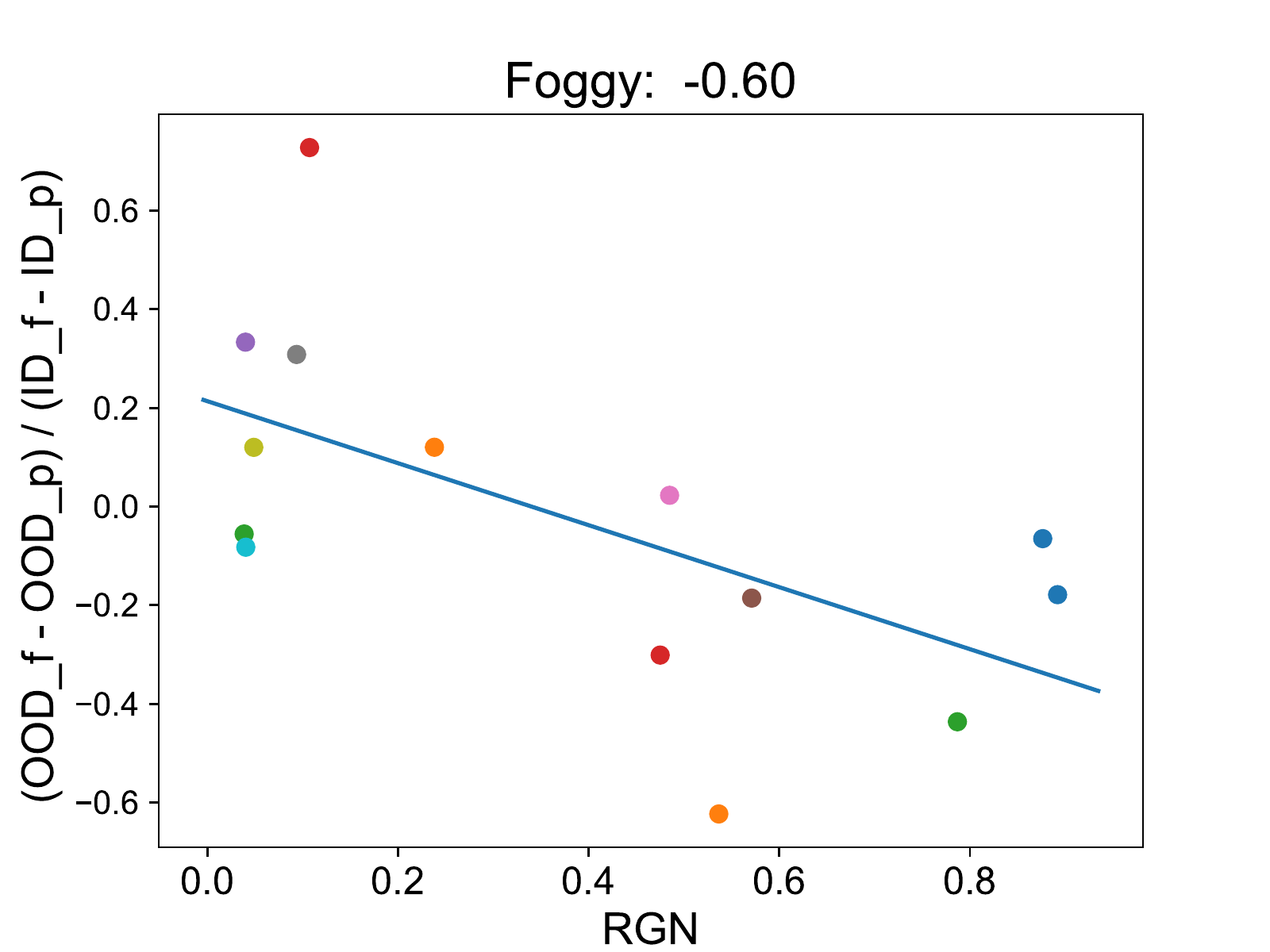}
  \caption{Foggy: -0.60 (mAP 10.9)}
  \label{fig:foggy}  
\end{subfigure}
\begin{subfigure}{0.33\textwidth}
  \centering
  \includegraphics[width=\linewidth]{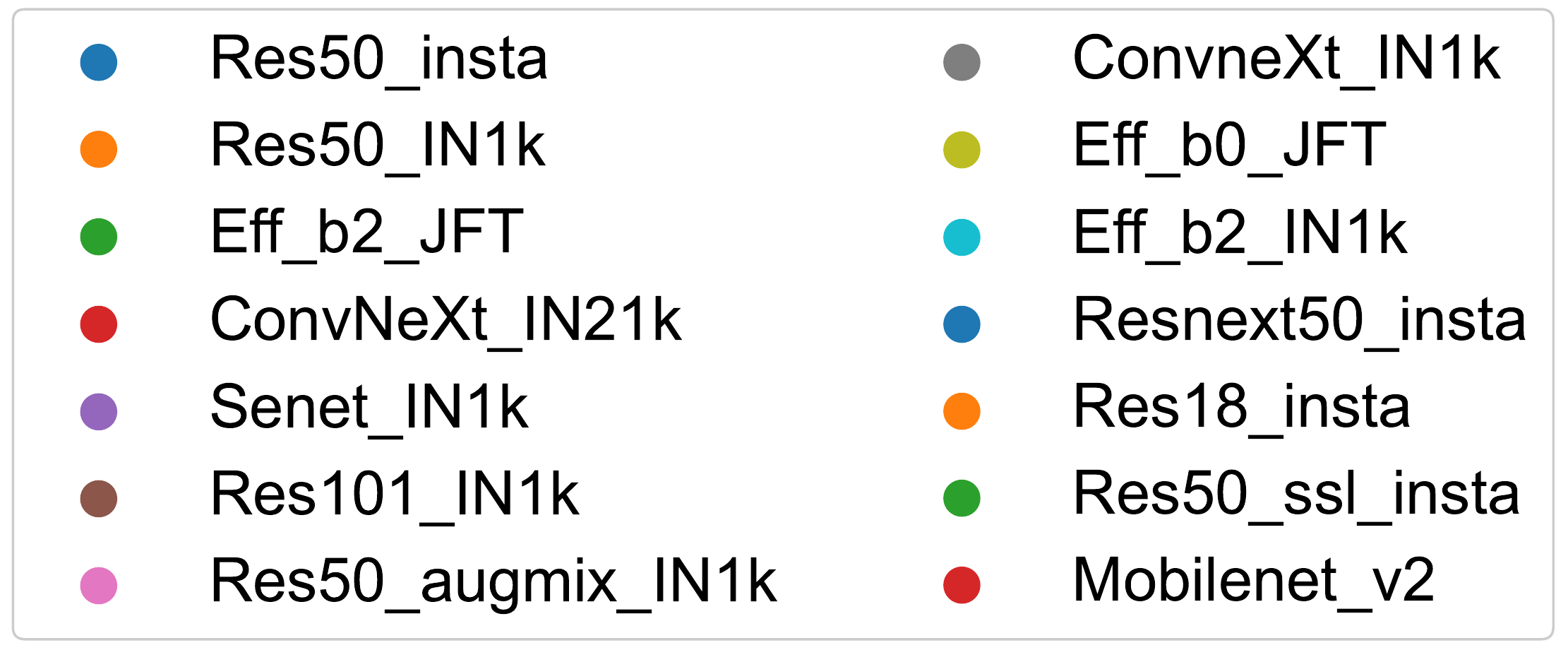}
      \vspace{+25pt}
\end{subfigure}
\vspace{-3mm}
\caption{A negative correlation exists between RGN and performance improvement in OOD over ID by fine-tuning. RGN (x-axis) indicates whether the network will degrade performance on OOD by fine-tuning. Y-axis indicates the ratio of mAP improvement in OOD to the improvement in ID by fine-tuning.}
\label{fig:rgn_vs_improvement}
\end{figure*}

%% file: tables/rgn_dataset_architecture.tex
\begin{table}[t]
\scalebox{0.9}{
\begin{tabular}{ccccc}

\toprule
Backbone    & Dataset     & SE    & RGN$\rm{_{pascal}}$ &RGN$\rm{_{city}}$ \\\hline
ResNet50        & IN1K      &  \xmark     & 0.31    & 0.53 \\
ResNet50        & IN1K + Augmix&  \xmark    & 0.57 &0.48   \\
ResNet50        & IN1K + Insta& \xmark  & 1.34  &0.89 \\ 
SeNet50         & IN1K       &  \cmark    & 0.05 &0.04   \\
\bottomrule
\end{tabular}
}
\vspace{-3mm}
\caption{Relationship among RGN and architecture and dataset used to train the model. RGN of SeNet is much smaller than that of ResNet. Note that the difference between SeNet50 and ResNet50 is whether \seblock is plugged into the model. Also, RGN is calculated except for \seblock for a fair comparison.} 
\label{tb:rgn_dataset_arch}
\end{table}


%% file: figures/seblock_histogram.tex
\begin{figure}
    \centering
    \includegraphics[width=\linewidth]{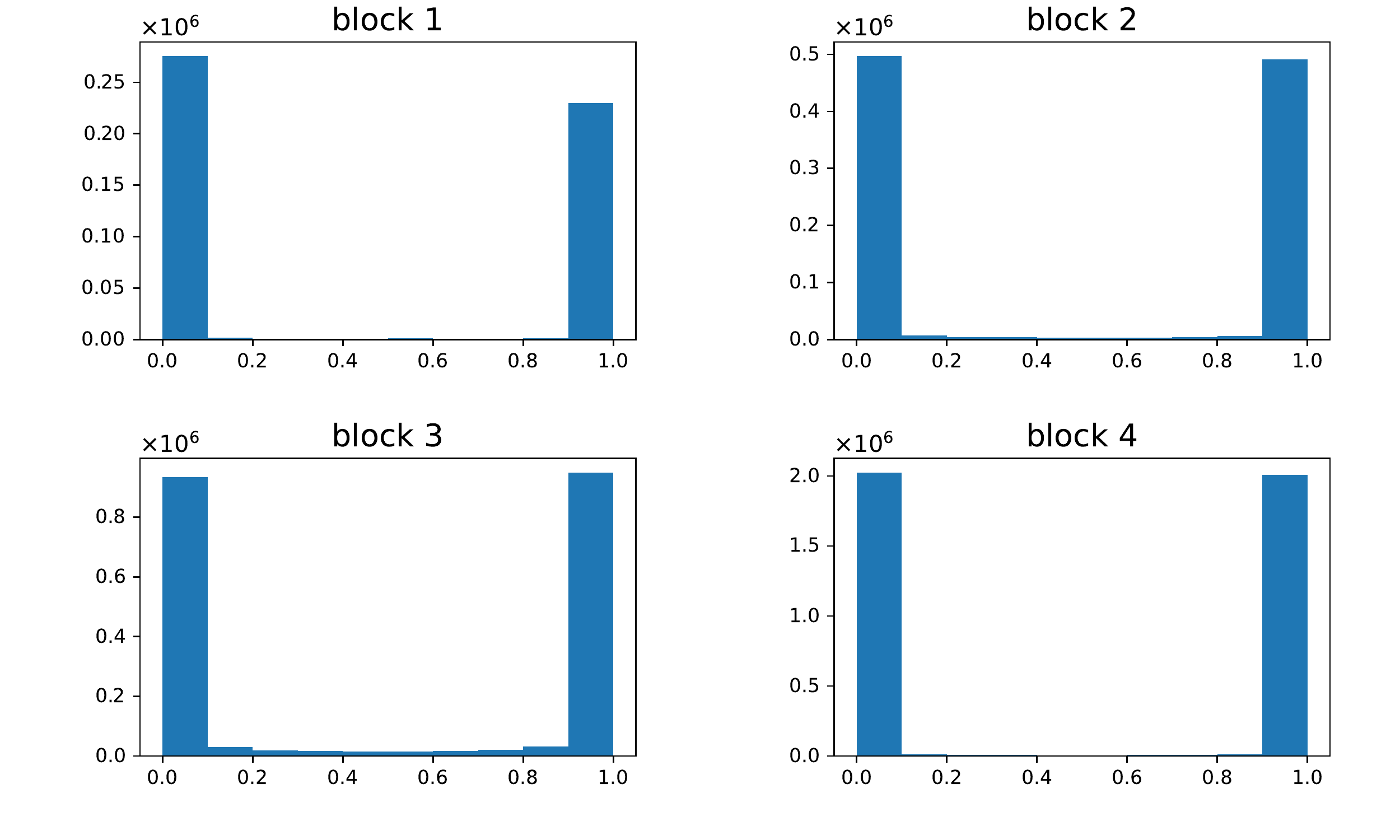}
    \vspace{-7mm}
    \caption{Visualization of \seblock activations. Note that most values are close to either zero or one, which works to suppress gradients from the upper layer.}
    \label{fig:se_histogram}
\end{figure}

%% file: figures/training_figure.tex
\begin{figure}
    \centering
    \includegraphics[width=\linewidth]{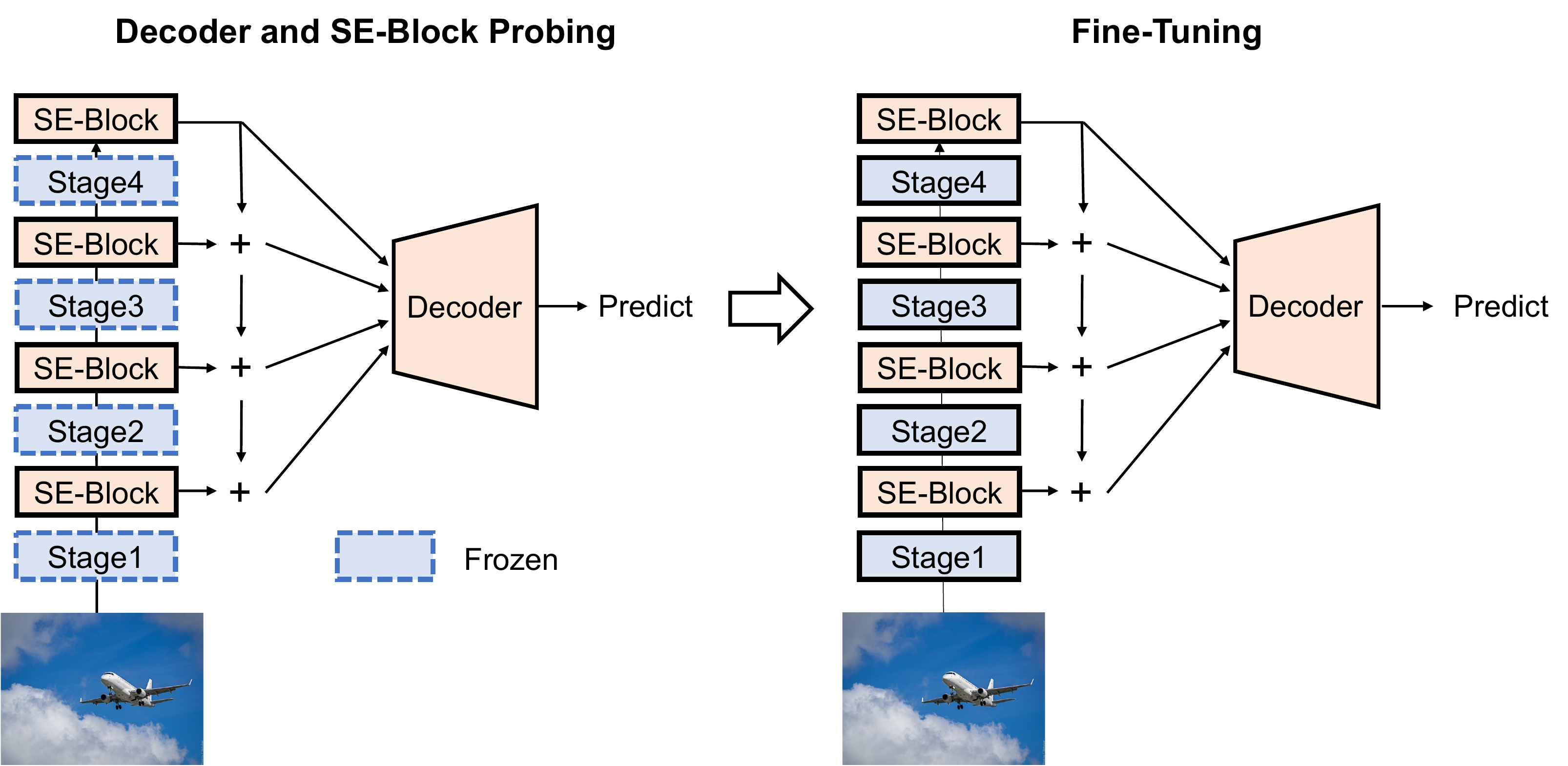}
    \vspace{-5mm}
    \caption{Illustration of decoder-probing with \seblock and (left) and its fine-tuning (right).}
    \label{fig:seblock_figure}
\end{figure}

%% file: tables/main_results.tex
\begin{table*}[t]
\centering
\begin{tabular}{c|c|c|c|c|c||c|c|c}
\toprule
\multirow{2}{*}{Backbone} &    \multirow{2}{*}{Method} & \multicolumn{4}{c||}{ID: Pascal} & \multicolumn{3}{c}{ID: Cityscapes}  \\
 &           & Comic & Watercolor & Clipart & ID                           & Foggy & BDD  & ID    \\\hline
\multirow{4}{*}{ResNet50 Instagram~\cite{mahajan2018exploring}}&DP                                   & 15.7  & 21.2       & 15.3    & 44.6 &                                    13.9 & 7.7 & 28.3  \\
&FT                                   & 7.5   & 19.4       & 11.4    & 50.4 &                                     12.8  & 5.1  & 33.5 \\
& \CCG DP-FT                                &\CCG 9.1   & \CCG21.0       & \CCG12.9    &\CCG 52.6 &\CCG14.8  &\CCG 5.5  &\CCG\bf{34.7}  \\
 &\CCG DP-FT + WR           &\CCG \bf{16.8}  &\CCG \bf{26.5}       &\CCG \bf{17.6}    &\CCG \bf{52.9} &\CCG \bf{19.3}  &\CCG \bf{9.6}  &\CCG 34.5   \\\hline
\multirow{4}{*}{ConvNeXt IN21K~\cite{liu2022convnet}}&DP                                   & 11.7  & 17.3       & 14.0    & 39.7 &                                     14.7 & 7.8 & 31.1 \\
&FT                              &      11.5  & 22.9       & 16.8    & 60.6 &  18.1 & 9.7 & 35.8    \\
&\CCG  DP-FT                   & \CCG              13.6  &\CCG  24.7       &\CCG  19.1    &\CCG  \bf{62.3} & \CCG                                   20.5&\CCG 11.5&\CCG   37.1\\
&\CCG DP-FT + WR     &\CCG       \bf{14.6}  &\CCG  \bf{27.8}       &\CCG  \bf{19.7}    &\CCG  61.4 & \CCG                          \textbf{21.1}          &\CCG    \textbf{11.7}     &\CCG  \bf{37.2}   \\\hline
\multirow{4}{*}{Eff-B2 JFT~\cite{xie2020self}}&  DP                                   & 12.6  & 20.4       & 15.1    & 40.2 &                                     11.1 & 6.9 & 25.2 \\
&FT                                   & 17.1  & 27.2       & 18.0    & 53.4 &          10.7                          &  5.1     &     31.5        \\
&\CCG  DP-FT                                &\CCG  17.4  &\CCG  29.4       &\CCG  20.7    &\CCG \bf{55.3} &\CCG   12.9&\CCG  7.3&\CCG   \bf{32.9}\\
&\CCG  DP-FT + WR        &\CCG  \bf{19.5}  &\CCG  \bf{30.0}         &\CCG  \bf{22.0}      &\CCG  54.2 &\CCG \bf{13.5} &\CCG  \bf{7.6} &\CCG  32.5\\

\bottomrule
\end{tabular}
\vspace{-3mm}
\caption{Effect of weight regularization. DP, FT, and WR denote decoder-probing, fine-tuning, and weight regularization.} 
\label{tb:main}
\end{table*}

%% file: paragraphs/experiments.tex
\section{Experiments}
\vspace{-1mm}
In this section, we first study the effect of weight regularization and \seblock. After that, we analyze the combination of the two techniques and other regularization, such as data augmentation. We follow the experimental protocol presented in Sec.~\ref{sec:analysis}. We set $\lambda$ in Eq.~\ref{eq:objective} as 0.1 based on the performance on Pascal to Watercolor using ResNet50, and apply the same value to all other settings with weight regularization. We analyze the sensitivity to $\lambda$ in the appendix.

\noindent\textbf{Effect of weight regularization.}
Table~\ref{tb:main} shows results with three pre-trained models. The results show the effectiveness of using weight regularization for all architectures. This means that even for the architecture with a robust structure, \eg, EfficientNet, adapting to downstream tasks without losing generalization is not trivial, thus, regularization is useful. Large $\lambda$ values in Eq.~\ref{eq:objective} impose strong regularization on the training, that is, larger $\lambda$ should improve OOD generalization while limiting the improvements on ID performance. We obtain empirical results consistent with this intuition in the appendix.

\input{tables/seblock_pascal}

\noindent\textbf{Inserting \seblock into the backbone.}
Table~\ref{tb:seblock} investigates the effectiveness of inserting \seblock into the pre-trained model following Sec.~\ref{sec:decoder}. Although it does not necessarily outperform the DP model in OOD, it consistently improves FT and DP-FT, showing that \seblock improves generalization. Additionally, combining the technique with weight regularization (see +WR) significantly improves performance, showing the compatibility of the two techniques.
Table~\ref{tb:seblock_rgn_analysis} compares DP and DP-SE models in terms of RGN and the detection loss computed on the checkpoints. 
The decrease in loss can reduce the RGN value, but even when the loss value increases a little (see ConvNeXt), DP-SE substantially reduces RGN value. This indicates that \seblock is important in preventing large gradients from passing to lower layers.

\noindent\textbf{Distance from the pre-trained model.}
Table~\ref{tb:weight_distance} shows the distance between fine-tuned (DP-FT) models and an initial model. Since weight regularization uses the distance as a penalty, the fine-tuned model gets closer to the initial one. Also, \seblock, the architecture level regularization, reduces the distance. Although the plain training renders the model far from the initial one, its performance on ID is worse than in some regularized models. Overfitting can cause performance degradation even ID. 

\input{tables/seblock_rgn_analysis}
\input{tables/distance_from_pretrained}

\input{figures/analysis_decoder}

\input{tables/compare_baselines.tex}

\noindent\textbf{Decoder producing small RGN helps OOD generalization.} 
Fig.~\ref{fig:long_train} compares DP, DP-FT, and DP-FT plus weight regularization by increasing the number of training iterations in the DP stage. We conduct training on Pascal with the ResNet50 model and report the performance averaged over the three OOD domains. 
We have two findings; (1) longer training in DP reduces RGN and improves OOD generalization in all approaches, yet, (2) the backbone still loses generalization by fine-tuning (DP vs. DP-FT), and (3) weight regularization is helpful to maintain the generalization. Fig.~\ref{fig:nas_fpn} further investigates the capacity of the decoder model using NAS-FPN~\cite{ghiasi2019fpn}, by increasing the stack of feature pyramid modules. Stacking more modules increases the generalization on OOD (NAS-FPN DP), but fine-tuning from such decoders significantly reduces generalization (NAS-FPN DP-FT). The decoder with more FPN stacks has a larger RGN, reducing generalization by fine-tuning. These models underperform a plain decoder with DP-FT trained with weight regularization (from Fig. ~\ref{fig:long_train}), showing that a stronger decoder does not necessarily produce a generalizable detector. We also try longer DP training for this strong decoder but do not see improvements in OOD performance because the strong decoder easily overfits the ID data. These observations suggest that training a lightweight decoder longer in the DP stage and applying weight regularization during fine-tuning is the best recipe to achieve an OOD robust detector.

\noindent\textbf{Compatibility with other DG approaches.}
Since the prior work on single-domain generalization for object detection~\cite{chen2021robust}, comparable to our setting, does not publish training code, we develop several DG approaches inspired by DG image classification to see the compatibility with our techniques in Table~\ref{tb:compare_baselines}.
First, to see the compatibility with data augmentation, we study the Augmix data augmentation~\cite{hendrycks2019augmix}. Comparing the second row (FT) and the last row (WR+SE), our approach and data augmentation have an orthogonal effect; data augmentation expands the training domain while our approach retains the knowledge of the pre-trained model. From this result, we conjecture that our approach is compatible with other augmentation, such as adversarial augmentation~\cite{chen2021robust}. 
Second, to ensure consistency with a pre-trained model, we introduce a model regularized with feature consistency loss. Specifically, given the DP model, this approach minimizes detection loss and contrastive loss between region-wise features from the DP and the tuned model, wherein Augmix~\cite{hendrycks2019augmix} is employed as data augmentation. Although the contrastive loss to ensure consistency with the pre-trained model is explored in image classification~\cite{chen2021contrastive}, our work is the first to test in object detection. The regularization approach, CR, outperforms FT trained with Augmix. Note that, thanks to the simplicity of weight regularization, it is easy to plug in, computationally efficient, and consistently improves the performance of CR (CR+WR). These results indicate that our proposed recipes are compatible with other generalization techniques.

\input{tables/swin_pascal}

\noindent\textbf{Transformer.} 
We further investigate the effectiveness of the regularization in Swin-Transformer~\cite{liu2021swin}. We employ Swin-B pre-trained with ImageNet-21K in Table~\ref{tb:swin_base}. While both FT and DP-FT decrease performance on OOD or the improvements are marginal, adding the weight regularization remarkably improves OOD generalization with a small decrease in ID performance. The advantage of using \seblock is not evident in this experiment. Further exploration of gradient-preserving layers in transformers is part of future work.

\input{tables/coco}
\noindent\textbf{Experiments on COCO.} 
We evaluate our approach using COCO~\cite{coco} and COCO-C~\cite{michaelis2019benchmarking} containing images with diverse types and severity of image corruptions. Table~\ref{tb:coco} shows that regularization boosts robustness to corruptions by more than five points over the plain fine-tuning (FT).

\input{tables/lvis}

\textbf{Analysis on the long-tailed instance segmentation.}
Although we focus on the input-level distribution shift in evaluation, one of the important shifts is in label distribution. Especially, long-tailed recognition aims to train a model which generalizes well on diverse categories from data with imbalanced label distribution.  
We hypothesize that the large-scale pre-trained backbone has representations effective at recognizing diverse categories, but it can lose the representations if trained on the imbalanced data; thus, the regularization on a backbone can be effective. Then, we conduct experiments on Lvis v1.0~\cite{gupta2019lvis} in Table~\ref{tb:lvis}.
We see that DP-FT degrades performance on rare categories (APr) compared to DP, while weight regularization improves the performance on all types of categories. This analysis indicates the effectiveness of a large-scale pre-trained model for long-tailed recognition.

%% file: tables/seblock_pascal.tex
\begin{table}[t]
\centering
\scalebox{0.9}{
\begin{tabular}{c|c|c|c|c|c}
\toprule
\multirow{2}{*}{Backbone} &    \multirow{2}{*}{Method} &\multicolumn{4}{c}{ID: Pascal}   \\
 &           & Comic & Water & Clipart & ID \\\hline
\multirow{5}{*}{\makecell{ResNet50 \\ Instagram}}&DP                                   & 15.7  & 21.2       & 15.3    & 44.6  \\
&FT                                   & 7.5   & 19.4       & 11.4    & 50.4 \\
           
& \CCG DP-FT                                &\CCG 9.1   & \CCG21.0       & \CCG12.9    &\CCG 52.6 \\
&\CCG DP-SE-FT                      &\CCG 10.2  & \CCG 22.5       &\CCG 15.1    &\CCG \bf{53.4} \\

&+ WR &\bf{18.9}  &\bf{27.5}       &\bf{21.4}    &52.2 \\\hline
\multirow{5}{*}{\makecell{ConvNeXt \\ IN21K}}&DP                                   & 11.7  & 17.3       & 14.0    & 39.7  \\
&FT                              &      11.5  & 22.9       & 16.8    & 60.6    \\
&\CCG  DP-FT                   & \CCG              13.6  &\CCG  24.7       &\CCG  19.1    &\CCG  \bf{62.3} \\
&\CCG  DP-SE-FT         &\CCG  15.4  &\CCG  27.5       &\CCG  20.9  &\CCG 61.6  \\
&+ WR              & \bf{17.8}  &  \bf{29.3}       & \bf{23.8}    &61.0  \\
\bottomrule
\end{tabular}
}
\vspace{-3mm}
\caption{Effect of inserting \seblock (SE). \seblock improves generalization and combining it with weight regularization (WR) in fine-tuning further increases performance. }
\label{tb:seblock}
\end{table}

%% file: tables/seblock_rgn_analysis.tex
\begin{table}[t]
\centering
\begin{tabular}{c|ccc}
\toprule
Backbone & Method&RGN & Loss \\ \hline
 \multirow{2}{*}{\makecell{ResNet50 \\ Instagram}}& DP& 1.34&0.34\\
 & DP-SE&0.36&0.27\\\hline
 \multirow{2}{*}{\makecell{ConvNeXt \\ IN21K}}& DP& 0.13&0.35\\
&DP-SE&0.04&0.36\\
\bottomrule
\end{tabular}
\vspace{-3mm}
\caption{Comparison between DP and DP-SE (decoder-probing inserting \seblock) in terms of RGN.}
\label{tb:seblock_rgn_analysis}
\end{table}

%% file: tables/distance_from_pretrained.tex
\begin{table}[t]
\centering
\scalebox{0.9}{
\begin{tabular}{cc|ccc}
\toprule
SE & WR & Distance from Init & AP (OOD) & AP (ID)\\\hline 
&&	7.91 & 14.3 & 52.6\\ 
\cmark&&6.41 & 15.9 &53.4\\
&\cmark&	0.34 & 19.9&52.9\\
\cmark&\cmark&	0.04 & 22.6&52.2\\
\bottomrule
\end{tabular}
}
\vspace{-3mm}
\caption{Distance from pre-trained models for DP-FT models. We use ResNet50 and Pascal. Both \seblock and weight regularization decrease the distance from the initial model, improving the generalization in OOD.}
\label{tb:weight_distance}

\end{table}

%% file: figures/analysis_decoder.tex
\begin{figure}[t]
\begin{subfigure}[b]{0.25\textwidth}
  \centering
  \includegraphics[width=1.1\textwidth]{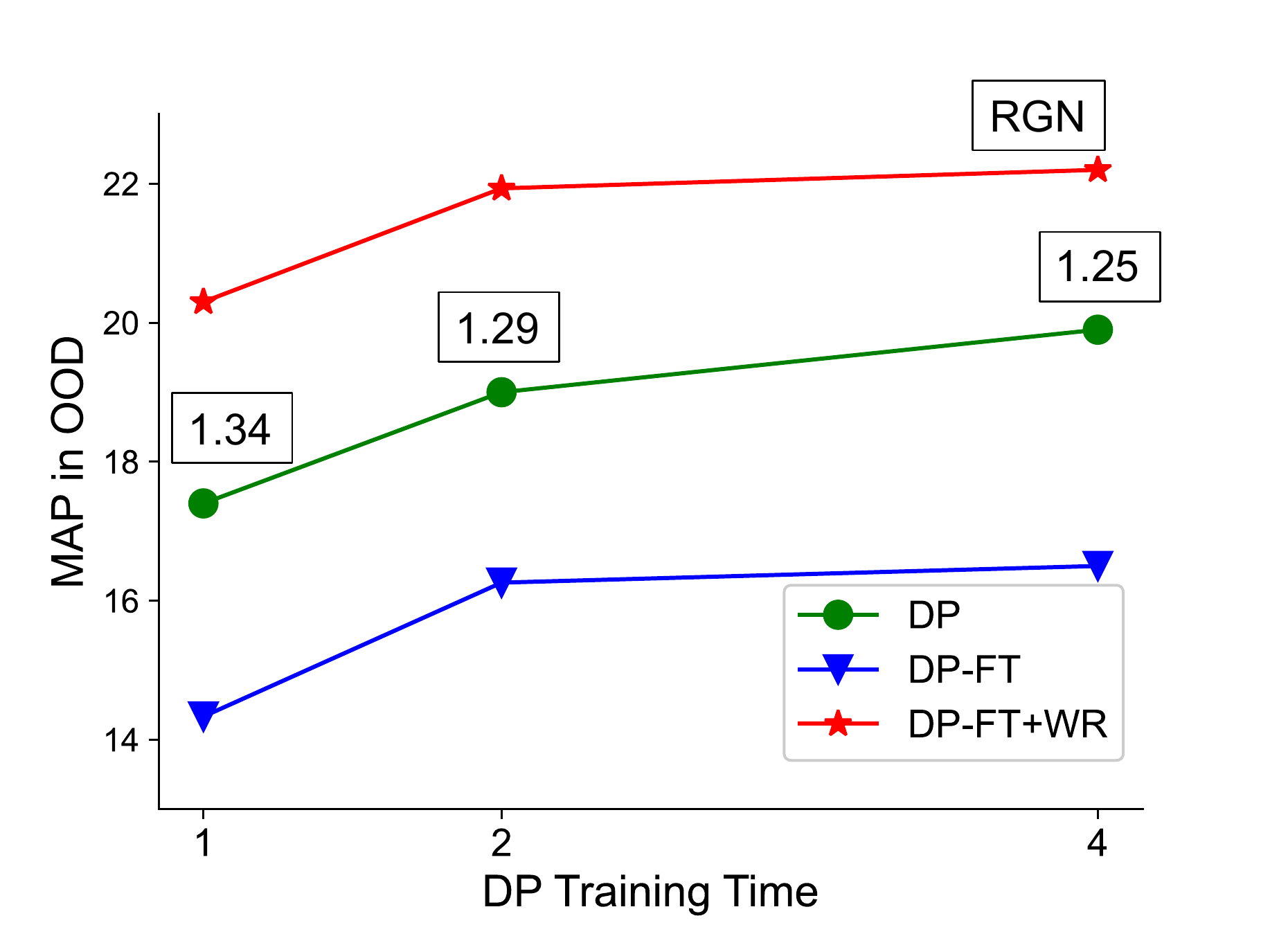}
  \caption{Longer DP training.}
  \label{fig:long_train}
\end{subfigure}
\begin{subfigure}[b]{.25\textwidth}
  \centering
  \includegraphics[width=1.1\textwidth]{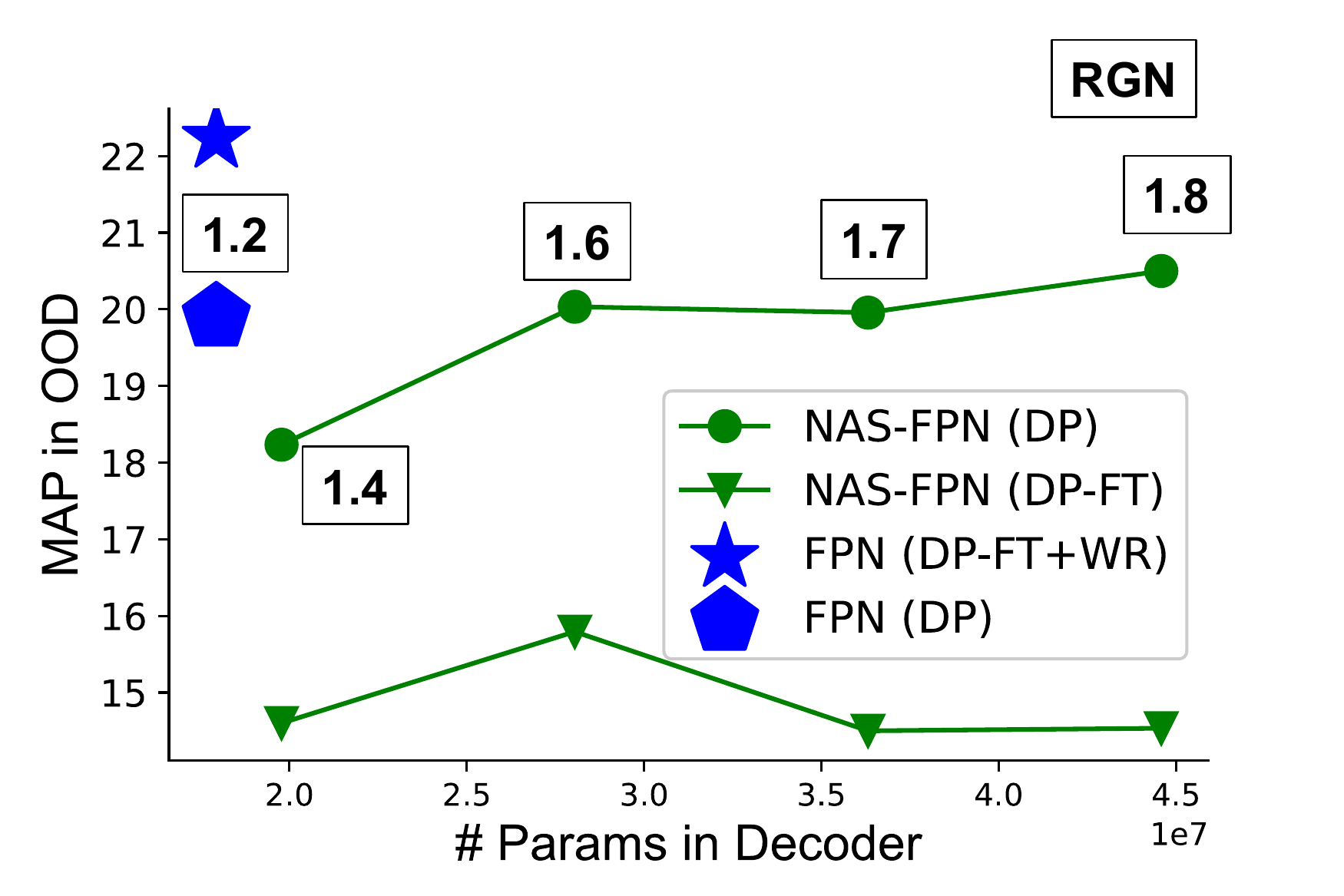}
  \caption{Stronger decoder.}
  \label{fig:nas_fpn}
\end{subfigure}
\vspace{-5mm}
\caption{Analysis of the decoder design. Left: longer DP training increases OOD performance. Right: Investigation on the stronger decoder (NAS-FPN~\cite{ghiasi2019fpn}). RGN value is shown next to each plot of DP. Strong decoder model increases RGN, which causes significant decrease in OOD performance in fine-tuning. }
\label{fig:analysis_decoder}
\end{figure}

%% file: tables/compare_baselines.tex
\begin{table*}[t]
\centering
\begin{tabular}{c|c|c|c|c|c|c||c|c|c}
\toprule
   \multirow{2}{*}{Method}  &  \multirow{2}{*}{DP-FT}  &\multirow{2}{*}{Augmix} &\multicolumn{4}{c||}{ID: Pascal} & \multicolumn{3}{c}{ID: Cityscape}  \\
            &&& Comic & Watercolor & Clipart & ID                           & Foggy & BDD  & ID    \\\hline
            FT& \xmark& \xmark&11.5&22.9&16.8&60.6&18.1&9.7&35.8\\  
            FT& \xmark & \cmark&\bf{15.5}&\bf{28.6}&\bf{20.7}&\bf{61.4}& \bf{25.4} & \bf{14.5} & \bf{37.6}\\\hline             
CR&\cmark&\cmark&17.5&31.1&23.3&\bf{59.9}&  26.9 & 14.3&  38.5\\

CR + WR &\cmark&\cmark&\bf{19.1}&\bf{32.2}&\bf{24.7}&58.6& \bf{27.8} & \bf{14.9} & \bf{39.2} \\\hline
WR + SE   &  \cmark       &\xmark &17.8&29.3&23.8&61.0 &21.5& 12.0 & 37.6  \\
WR + SE&\cmark& \cmark&\bf{20.0}&\bf{32.9}&\bf{25.4}& \bf{61.3}& \bf{27.1} & \bf{14.8} & \bf{38.8} \\
\bottomrule
\end{tabular}
\vspace{-3mm}
\caption{Compatibility of our weight regularization (WR) and \seblock training (SE) techniques with other domain generalization techniques using ConvNeXt. CR: contrastive regularization, Augmix: data augmentation.}
\label{tb:compare_baselines}
\end{table*}

%% file: tables/swin_pascal.tex
\begin{table}[t]
\centering
\scalebox{0.95}{
\begin{tabular}{c|c|c|c|c}
\toprule
\multirow{2}{*}{Method} &\multicolumn{4}{c}{ID: Pascal}   \\
            & Comic & Water & Clipart & ID \\\hline
DP &  17.9&27.2&22.6&49.5\\
FT &13.7&29.0&23.1&62.3\\
\CCG DP-FT &\CCG 15.4&\CCG 30.7&\CCG 23.2&\CCG  \bf{62.8}\\
\CCG DP-FT + WR &\CCG \bf{19.5}&\CCG  \bf{33.7}&\CCG  \bf{26.3}&\CCG 61.9\\
\CCG DP-SE-FT + WR & \CCG \bf{19.8}&\CCG \bf{33.1}&\CCG \bf{26.8}&\CCG 60.6\\
\bottomrule
\end{tabular}
}
\vspace{-3mm}
\caption{Results on Swin Transformer. }
\label{tb:swin_base}
\end{table}

%% file: tables/coco.tex
\begin{table}[t]
\centering
\scalebox{0.9}{
\begin{tabular}{ccc|c|c|c}
\toprule
   DP-FT&WR&SE&COCO& COCO-C& rPC (\%)  \\\hline
   &&& 36.5 &20.2&55.3\\ 
   \cmark  &&& \textbf{38.5}& 22.7& 58.9\\   
   \cmark  &\cmark&&37.7& 25.1&  66.5 \\
   \cmark  &\cmark&\cmark&  37.2&\textbf{25.7}&  \textbf{69.0}\\
\bottomrule
\end{tabular}
}
\vspace{-3mm}
\caption{Results of ResNet50 trained on COCO (AP). rPC denotes the
relative performance under corruption. Introducing weight regularization and \seblock substantially improves robustness over FT (the first row).}
\label{tb:coco}
\end{table}

%% file: tables/lvis.tex
\begin{table}[t]
\centering
\scalebox{0.8}{
\begin{tabular}{c|cccc|cccc}
\toprule
\multirow{2}{*}{Method} &\multicolumn{4}{c|}{Bbox} &\multicolumn{4}{c}{Seg}  \\
            &  AP & APr & APc & APf & AP & APr & APc & APf \\\hline
DP &  20.1&10.2&19.2&25.5&20.9&11.8&20.4&25.5\\
DP-FT &22.1 & 9.9&20.3&\textbf{29.5}&22.2&11.4&20.9&28.4\\ 
DP-FT + WR &\textbf{22.4}& \textbf{10.7}& \textbf{21.1}&29.0&\textbf{22.9}&\textbf{11.9}&\textbf{22.3}&\textbf{28.5}\\
\bottomrule
\end{tabular}
}
\caption{Results on long-tailed instance segmentation using Lvis v1.0~\cite{gupta2019lvis}. }
\label{tb:lvis}
\end{table}

%% file: paragraphs/conclusion.tex
\section{Conclusion}
We study ways of achieving a robust object detector by fine-tuning from a robust pre-trained model. 
We explore  warmup fine-tuning in object detection, i.e., decoder-probing followed by full fine-tuning, and find that it gives architecture-specific improvements. Given this, we explore two complementary techniques to preserve  generalizable representations:  weight regularization to preserve backbone features, and the design of the decoder. These techniques prove useful for diverse architectures and datasets. We believe they will be good baselines for generalizable object detection. 
\section{Acknowledgement}
This work was supported by DARPA LwLL.

%% file: paragraphs/supple.tex
\section{Experimental Details}
\noindent \textbf{Datasets.}
The number of test images used for evaluation is listed in Table~\ref{tb:dataset_summary}. For experiments on COCO, we use the validation split for evaluation. 

\input{tables/dataset_summary}
\noindent \textbf{Training Details.}
We set up the hyper-parameters following the instructions provided by detectron2~\cite{wu2019detectron2}, \eg, we train the model for 80,000 iterations with a batch size of 2 in Pascal and 24,000 iterations with a batch size of 8 in Cityscape. In Cityscape, MaskRCNN is trained and evaluation is done on the instance segmentation task. 
For other architectural choices \eg, the architecture of the feature pyramid and detector head, we employ the default configuration. 
One exception is in the training of ConvNeXt~\cite{liu2022convnet}, where we use the group normalization~\cite{wu2018group} in the feature pyramid module to stabilize the training. 
For pre-trained models, we employ weights available in PyTorch Image Models~\cite{rw2019timm}.
We will publish the code used for training, including each configuration, upon acceptance. 

\section{Additional Results}
\input{tables/cityscape}

\textbf{Results on Cityscape.}
Table~\ref{tb:cityscape} shows the full ablation study on Cityscape using ConvNeXt. Overall, combining all modules performs well in many domains.

\input{tables/dataset_comparison}
\textbf{Analysis of the dataset used for pre-training.} 
Table~\ref{tb:compare_dataset} describes the performance of ResNet50 models pre-trained with different datasets or data augmentation, where we apply weight regularization to train these models. Generally, pre-training on diverse data makes the model generalize well on OOD datasets. Also, improvements over the FT baseline by the regularization get more significant in the pre-training. In other words, pre-trained on diverse data, the model will likely forget the learned representations with standard training.

\input{figures/lamda.tex}
\input{figures/compare_regularization}
\input{figures/nas_fpn_analysis}

\textbf{ID-OOD Trade-off by the coefficient.}
Fig.~\ref{fig:trade-off} shows a trade-off in ID-and-OOD performance controlled by the regularization coefficient, $\lambda$ in Eq. 4, which is measured using Pascal with ResNet50-Instagram. Increasing $\lambda$ adds more regularization to keep the parameter near the initial model, reflected as the performance decrease in ID (Pascal). Clipart and Comic increase the performance with more regularization, while Watercolor peaks near 0.1. The peak of OOD performance should differ by the similarity with ID dataset.

\input{tables/compare_regularization}

\textbf{Comparison among different regularizations.}
Table~\ref{tb:compare_regularization} compares different regularization, \ie, weight regularization with L2-distance, EWC, and RGN weighted regularization, using DP-FT. 
We train models using Pascal and pick $\lambda$ based on the Watercolor domain and show the average over three domains as OOD. We do not see a significant difference among the three regularizations, in particular, in OOD performance. We further investigate the strength of the regularization on each $\lambda$ in Fig.~\ref{fig:compare_reg}. The Y-axis shows the MAP on Pascal with each coefficient over the MAP of the DP-FT model. With more regularization, the value decreases. Therefore, the value implies the strength of the regularization. Then, we see that the strength of the regularizer is similar across different architectures by RGN while the other two show different strengths. This indicates that RGN can add similar strength of the regularization on different architectures by the same $\lambda$. This is because the regularization is weighted by the relative gradient norm, which should adjust the strength of the regularization depending on the norm of the gradient and parameter. In short, L2-distance and EWC may require additional hyper-parameter $\lambda$ tuning when the architecture is changed while RGN weighted regularization is less sensitive to the changes in the architecture.
\input{tables/robustness}

\textbf{Analysis on NAS-FPN.}
Fig.~\ref{fig:nas_fpn_dp_long} shows the effect of training iterations for the DP model of NAS-FPN and FPN, where we set the number of stacks in NAS-FPN as 3. The NAS-FPN shows strong performance on ID by longer training but reduces the performance on OOD. By contrast, FPN improves the performance on both types of distribution through longer training. A strong decoder can easily overfit ID data; thus, OOD performance can decrease with longer training. 

\input{figures/robustness}
\textbf{Analysis on the robustness to image corruptions.}
Table~\ref{tb:robustness_details} shows the results on the robustness to image corruptions using COCO. Using our regularization, the trained detector improves the robustness to all types of corruption. Interestingly, the use of \seblock decreases the robustness to the high-frequecy noise, \eg, gaussian noise, while improving the robustness to the other types of corruptions. The architectural difference seems to cause this change. 
Fig.~\ref{fig:severity} shows the performance on each severity of the corruptions. Trained with regularization (WR, SE), the detector performs better on the diverse level of severities than the model without regularization. 


\input{figures/ex_pascal}
\input{figures/ex_cityscape}
\textbf{Qualitative Results.}
Fig.~\ref{fig:ex_pascal} and ~\ref{fig:ex_cityscape} show qualitative results on Pascal and Cityscape. Note that, using our regularization, more objects are detected, \eg, middle in Fig.~\ref{fig:ex_pascal} and top right in Fig.~\ref{fig:ex_cityscape}, or correctly classified, \eg, top left in Fig.~\ref{fig:ex_pascal}. Several objects are not localized by both models, which indicates the difficulty of this task.

%% file: tables/dataset_summary.tex
\begin{table}[h]
\centering
\begin{tabular}{c|c|c|c|c|c|c}
\toprule
  \multicolumn{4}{c|}{ID: Pascal} & \multicolumn{3}{c}{ID: Cityscape}  \\
Comic & Water & Clipart & ID  & Foggy & BDD  & ID    \\\hline
1,000&1,000&500&1,999&500&5,346&500\\            
\bottomrule
\end{tabular}
\caption{Number of test images.}
\label{tb:dataset_summary}
\end{table}

%% file: tables/cityscape.tex
\begin{table}[h]
\centering
\begin{tabular}{c|cccc}
\toprule
Method & Foggy & BDD  & Cityscape    \\\hline
DP&14.7 & 7.8 & 31.1 \\
FT    &  18.1 & 9.7 & 35.8    \\
\CCG  DP-FT & \CCG20.5&\CCG 11.5&\CCG  37.1\\
\CCG  DP-SE-FT         &\CCG   \bf{22.0}  &\CCG  11.3 &\CCG  36.6            \\
\CCG DP-FT + WR     &\CCG           21.1                         &\CCG   11.7     &\CCG  \bf{37.2}             \\
\CCG  DP-SE-FT + WR             &\CCG  21.7&\CCG  \bf{11.8}&\CCG  \bf{37.2}\\
\bottomrule
\end{tabular}
\caption{Results on Cityscapes using Convnext.}
\label{tb:cityscape}
\end{table}

%% file: tables/dataset_comparison.tex
\begin{table}[t]
\centering
\begin{tabular}{c|c|c|c}
\toprule
   \multirow{2}{*}{Data} &\multicolumn{3}{c}{ID: Pascal}  \\\cline{2-4}
&            Comic & Water & Cart \\\hline 
IN1K&10.9 (\textcolor{blue}{+3.6})&22.5 (\textcolor{blue}{+2.1}) &16.5 (\textcolor{blue}{+4.7})\\
IN1K+Augmix&    12.7 (\textcolor{blue}{+3.1}) &25.7 (\textcolor{blue}{+2.6}) &17.3 (\textcolor{blue}{+2.5}) \\
Instagram&16.8	(\textcolor{blue}{+9.3}) & 26.5	(\textcolor{blue}{+7.1}) &17.6 (\textcolor{blue}{+6.2})   \\
\bottomrule
\end{tabular}
\caption{Results of ResNet50 pre-trained with different datasets or augmentation. We show the results with DP-FT + WR and improvement over FT baseline with blue numbers.}
\label{tb:compare_dataset}
\end{table}

%% file: figures/lamda.tex
\begin{figure}[ht]
    \centering
    \includegraphics[width=0.8\linewidth]{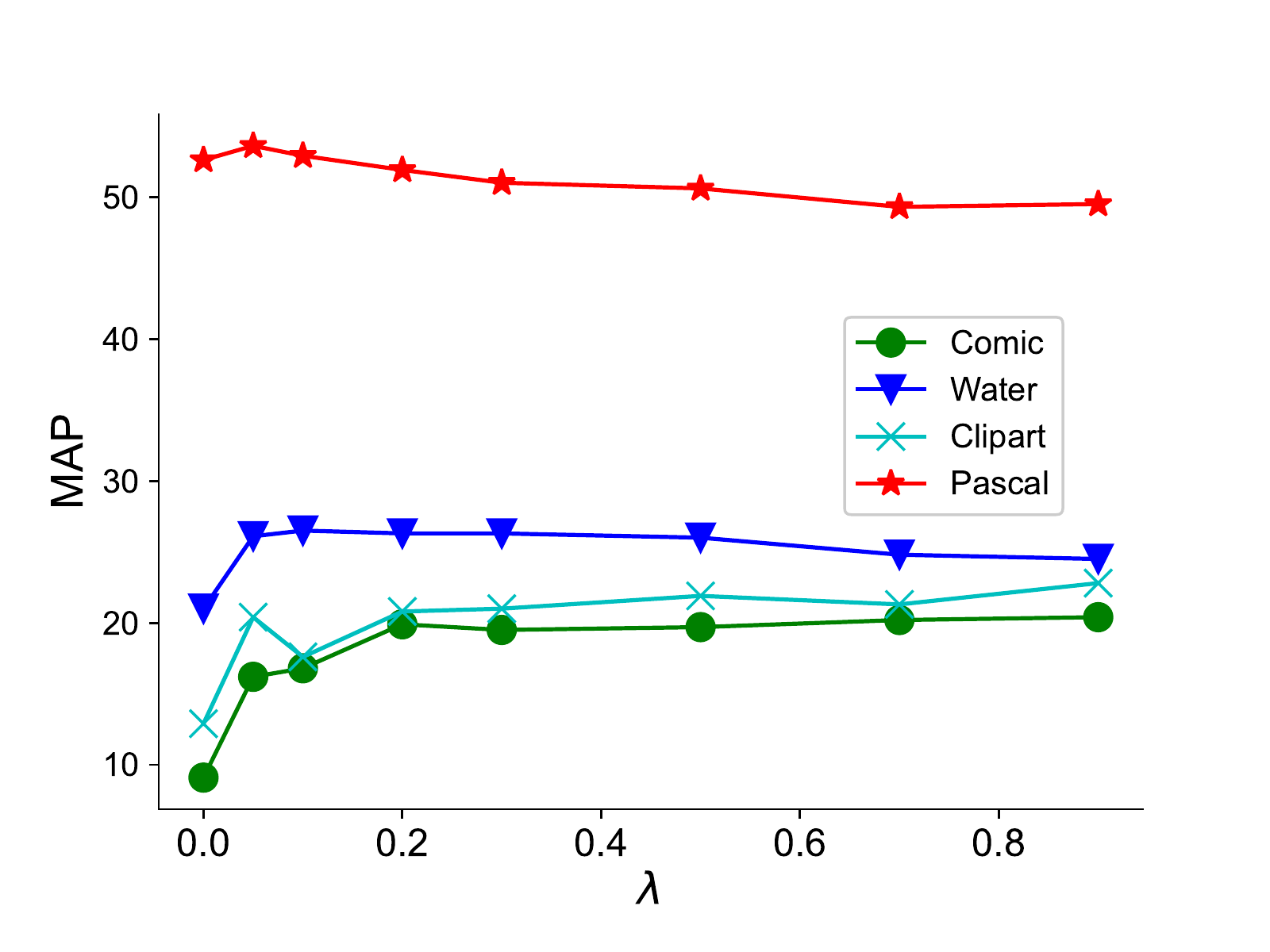}
    \caption{Sensitivity to the hyper-parameter.}
    \label{fig:trade-off}
\end{figure}

%% file: figures/compare_regularization.tex
\begin{figure*}[t]
\centering
\begin{subfigure}[b]{0.3\textwidth}
  \centering
  \includegraphics[width=\textwidth]{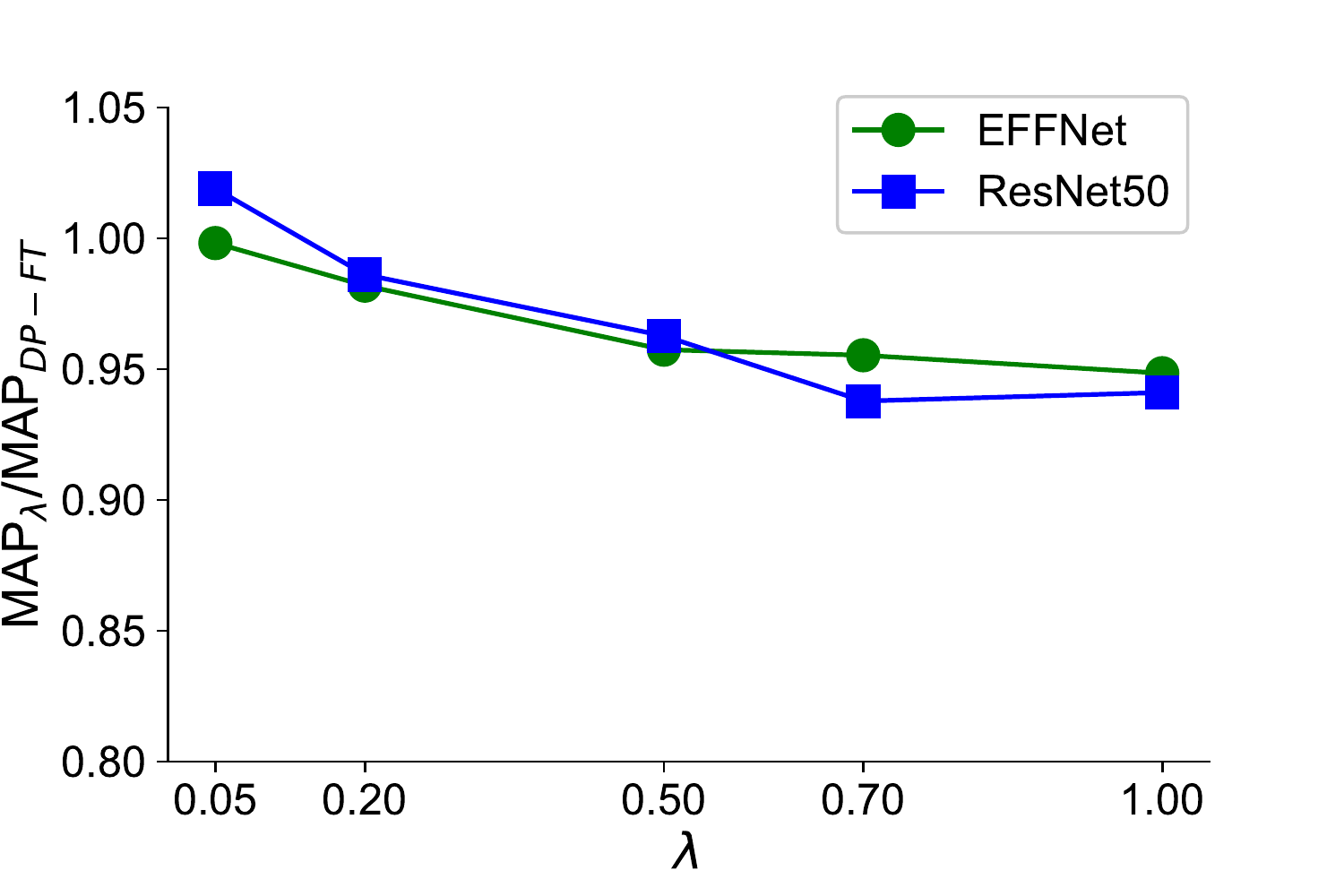}
  \caption{RGN weighted regularization}
  \label{fig:rgn_reg}
\end{subfigure}
\begin{subfigure}[b]{.3\textwidth}
  \centering
  \includegraphics[width=\textwidth]{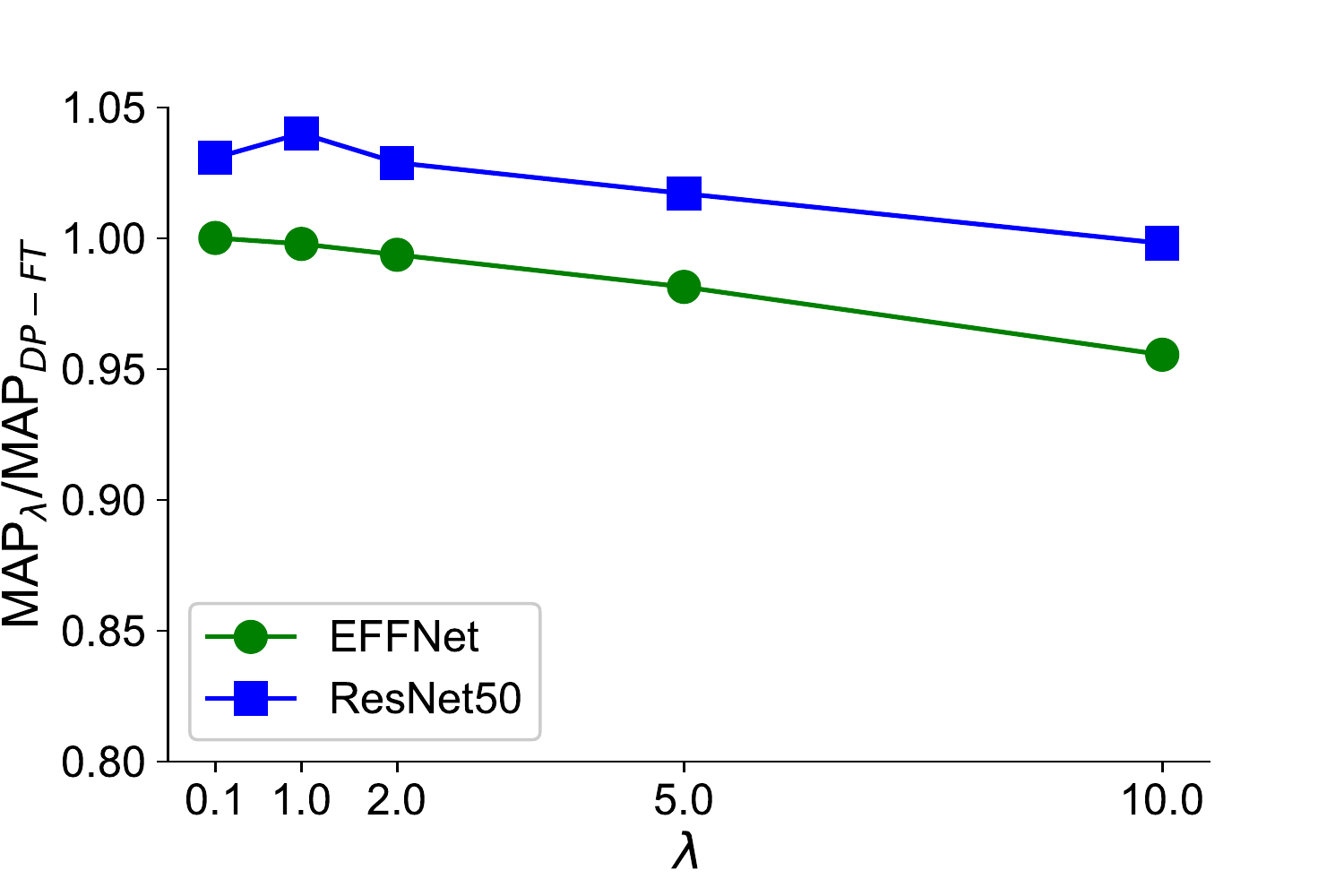}
  \caption{EWC}
  \label{fig:ewc_reg}
\end{subfigure}
\begin{subfigure}[b]{0.3\textwidth}
  \centering
  \includegraphics[width=\textwidth]{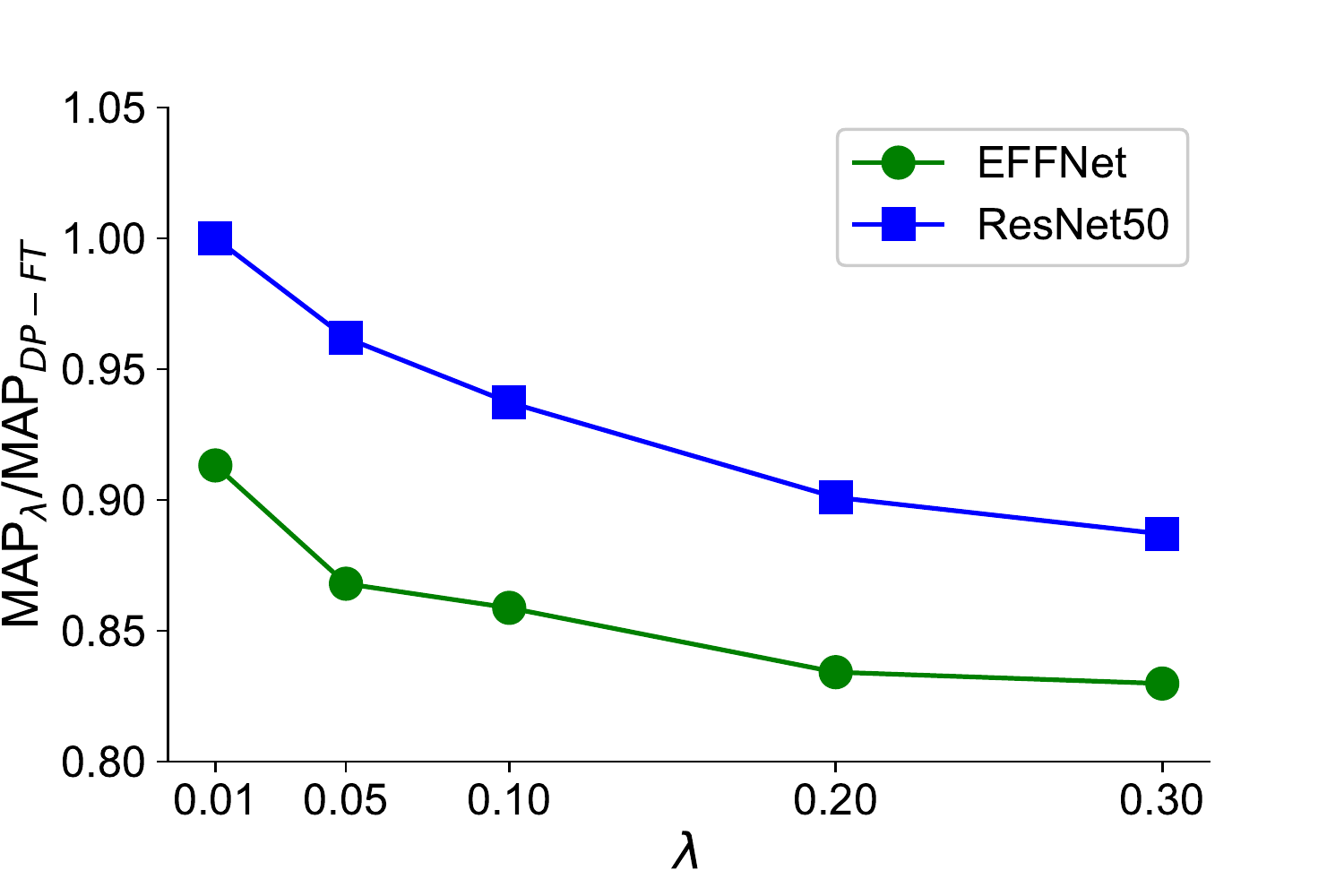}
  \caption{L2 distance}
  \label{fig:plain_reg}
\end{subfigure}
\caption{Comparison among difference regularization. (a): RGN weighted regularization. (b): EWC. (c): Plain weight regularization. X-axis varies the coefficient for regularization. Y-axis denotes the MAP on the training domain (Pascal) with each coefficient over the MAP of the DP-FT model. Intuitively, Y-axis denotes the strongness of the regularization.}
\label{fig:compare_reg}
\end{figure*}

%% file: figures/nas_fpn_analysis.tex
\begin{figure*}[t]
\begin{subfigure}[b]{0.25\textwidth}
  \centering
  \includegraphics[width=1.1\textwidth]{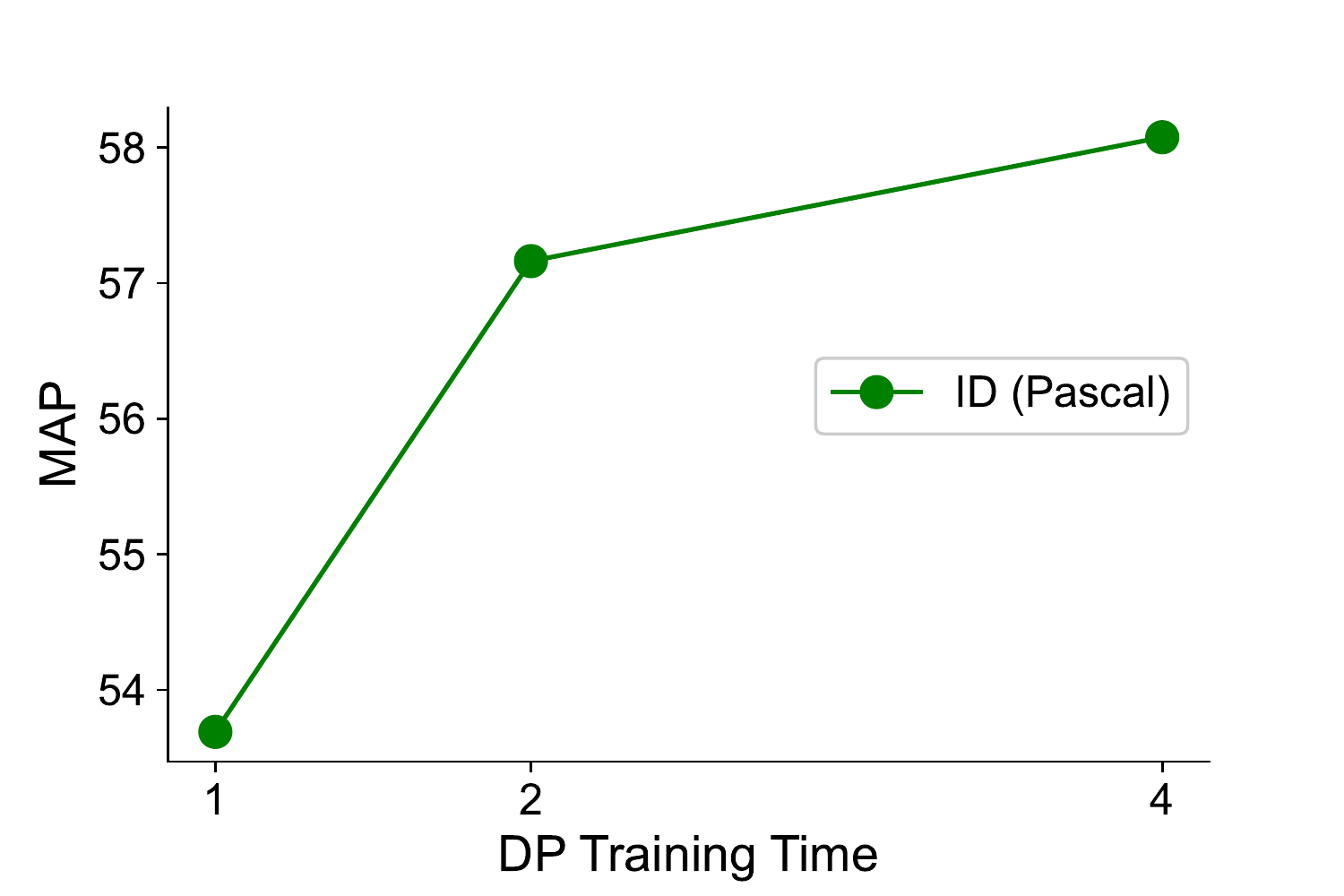}
  \caption{MAP on ID (Pascal).}
  \label{fig:nas_fpn_id}
\end{subfigure}
\begin{subfigure}[b]{.25\textwidth}
  \centering
  \includegraphics[width=1.1\textwidth]{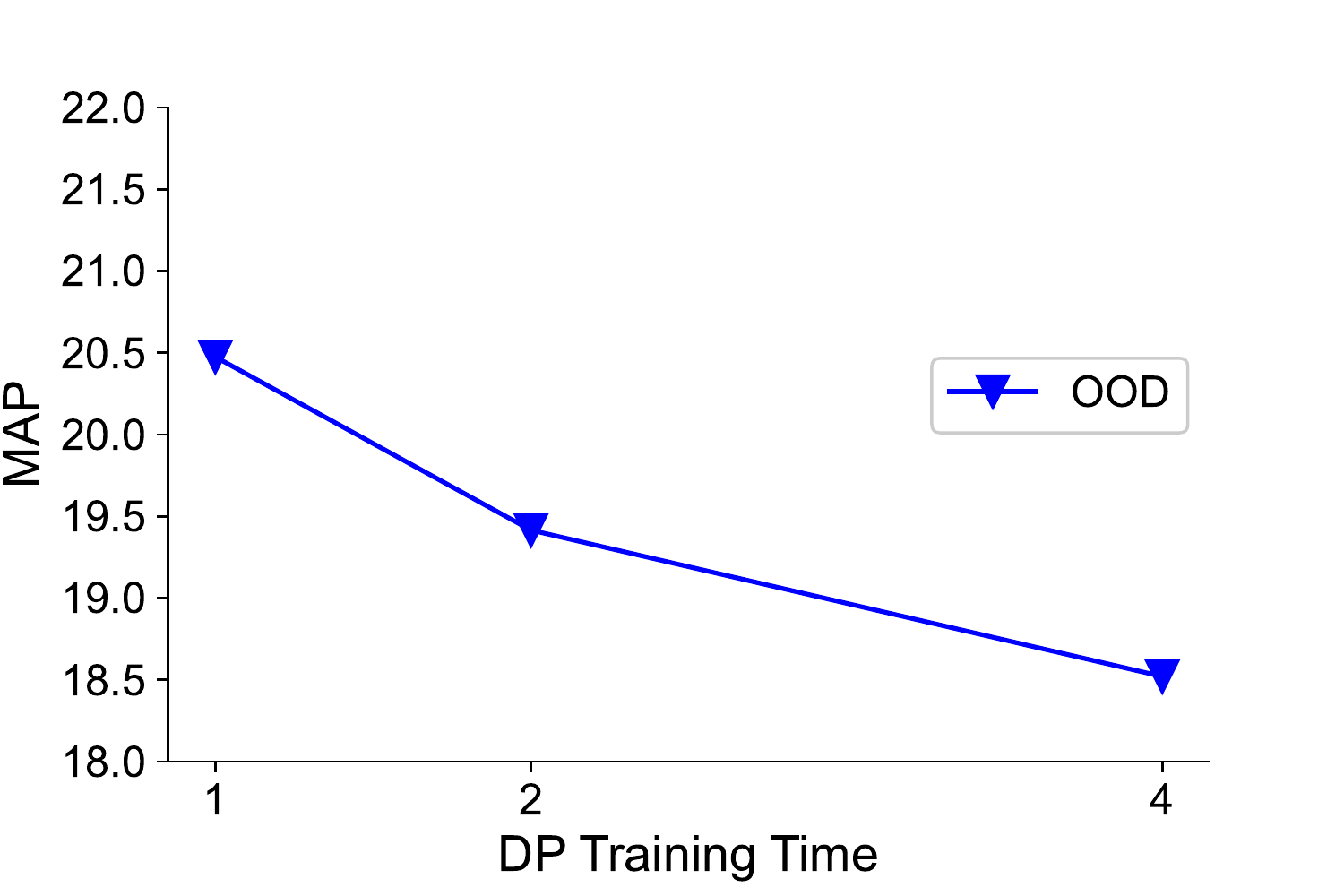}
  \caption{MAP on OOD.}
  \label{fig:nas_fpn_ood}
\end{subfigure}
\begin{subfigure}[b]{0.25\textwidth}
  \centering
  \includegraphics[width=1.1\textwidth]{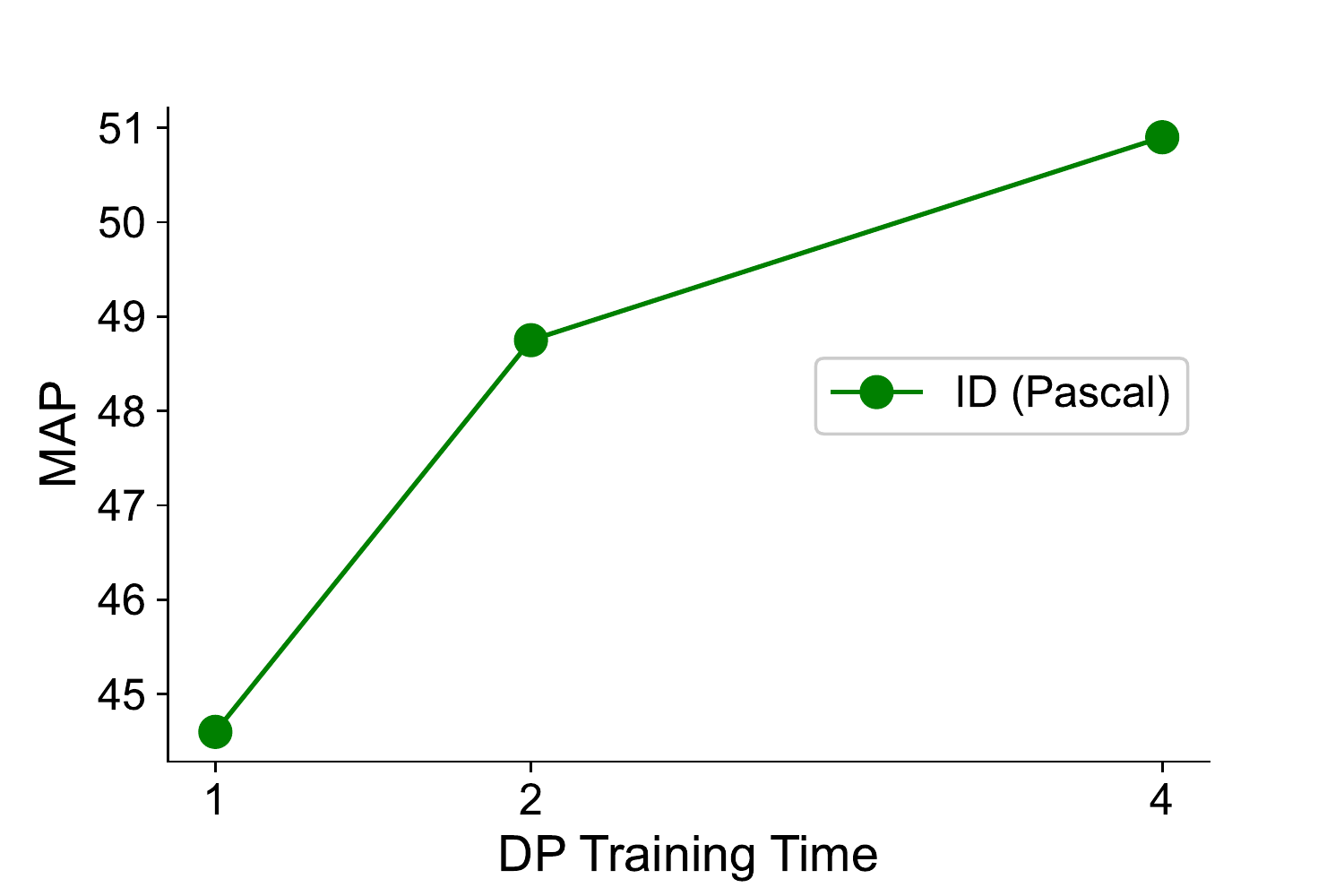}
  \caption{MAP on ID (Pascal).}
  \label{fig:fpn_id}
\end{subfigure}
\begin{subfigure}[b]{.25\textwidth}
  \centering
  \includegraphics[width=1.1\textwidth]{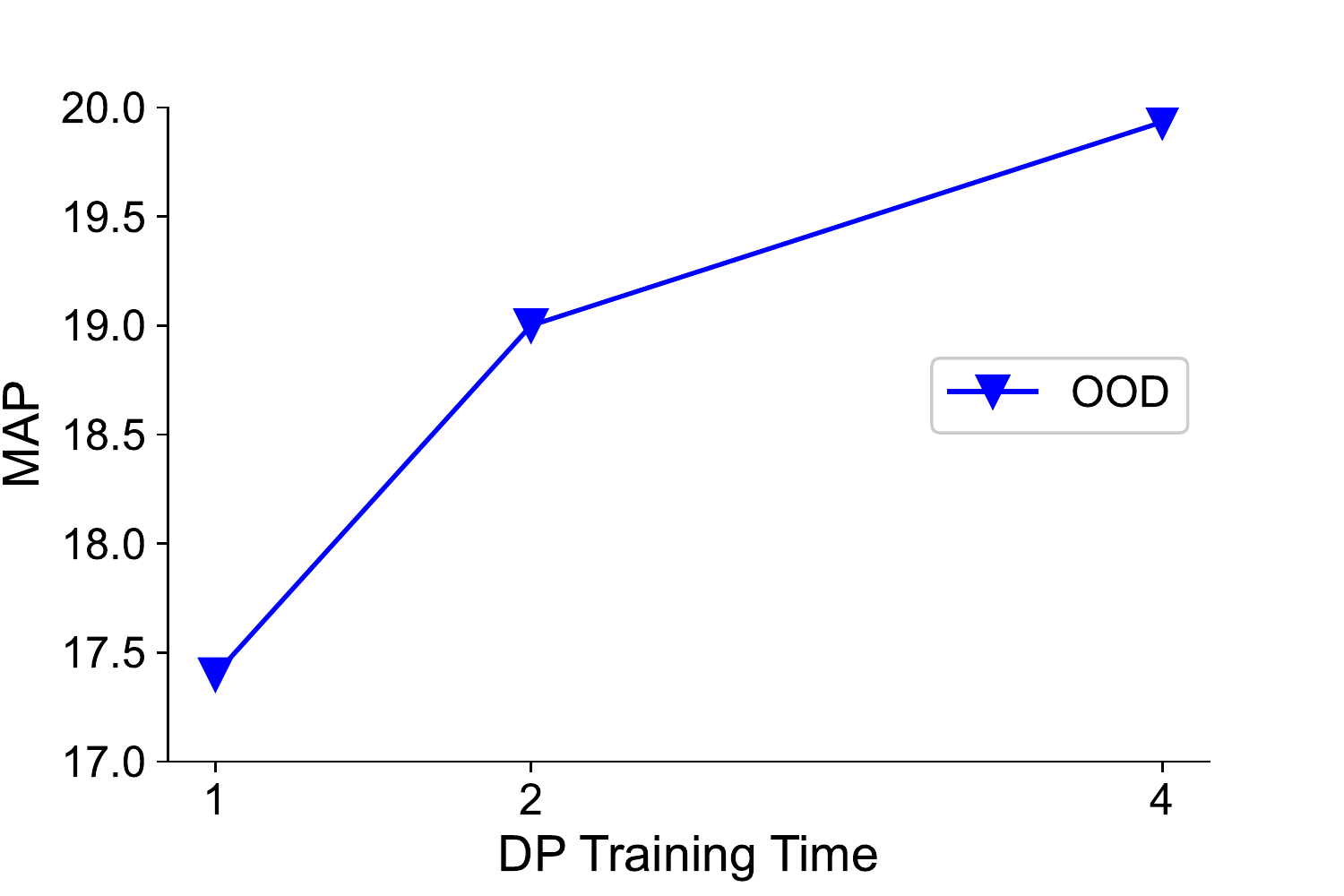}
  \caption{MAP on OOD.}
  \label{fig:fpn_ood}
\end{subfigure}
\caption{Effect of longer training in DP with NAS-FPN (a)(b) and the default FPN (c)(d).}
\label{fig:nas_fpn_dp_long}
\end{figure*}

%% file: tables/compare_regularization.tex
\setlength{\tabcolsep}{5pt}

\begin{table}[t]
\centering
\begin{tabular}{c|cc|cc}
\toprule
   \multirow{2}{*}{Data} &\multicolumn{2}{c|}{EfficientNet} &\multicolumn{2}{c}{ResNet50}  \\\cline{2-5}
&   OOD & ID & OOD&ID \\\hline 
L2-penalty&22.3&50.5&20.5&52.6\\ 
EWC &23.3&55.2&21.0&53.5\\ 
RGN&23.8&54.3&20.3&52.9\\
\bottomrule
\end{tabular}
\caption{Performance comparison among different regularization.}
\label{tb:compare_regularization}
\end{table}

%% file: tables/robustness.tex
\setlength{\tabcolsep}{1.5pt}
\begin{table*}[t]
\centering
\begin{tabular}{ccc|c|ccc|cccc|cccc|cccc}
\toprule

   DP-FT&WR&SE&Clean&\small{Gauss.}&\small{Shot}&\small{Impulse}&\small{Defocus}&\small{Glass}&\small{Motion}&\small{Zoom}& \small{Snow}&\small{Frost}&\small{Fog}&\small{Bright}&\small{Contrast}&\small{Elastic}&\small{Pixel}&\small{JPEG} \\\hline
   &&& 36.5 &18.5&19.2&16.8&24.9&17.4&21.2&6.4&12.8&19.1&20.8&	31.9&21.1&28.0&23.6&20.5\\ 
   \cmark  &&& \textbf{38.5}&21.0&21.8&19.2&26.9&18.7&22.7&6.7&	14.2&21.7&32.1&33.9&23.7&29.8&26.0&21.9\\   
   \cmark  &\cmark&&37.7&\bf{23.4}&\bf{23.8}&\bf{21.7}&27.1&20.7&24.2&7.9&16.6&	23.0&34.1&33.8&27.8&30.4&32.8&28.6\\
   \cmark  &\cmark&\cmark&  37.2&22.7&23.0&20.8&\bf{28.6}&\bf{23.0}&\bf{25.4}&\bf{8.1}	&\bf{17.4}&\bf{23.5}&\bf{34.5}&\bf{34.2}&\bf{28.7}&\bf{30.7}&\bf{33.2}&\bf{31.0}\\
\bottomrule
\end{tabular}
\caption{Detailed results on the robustness evaluation using COCO.}
\label{tb:robustness_details}
\end{table*}

%% file: figures/robustness.tex
\begin{figure}
    \centering
    \includegraphics[width=0.8\linewidth]{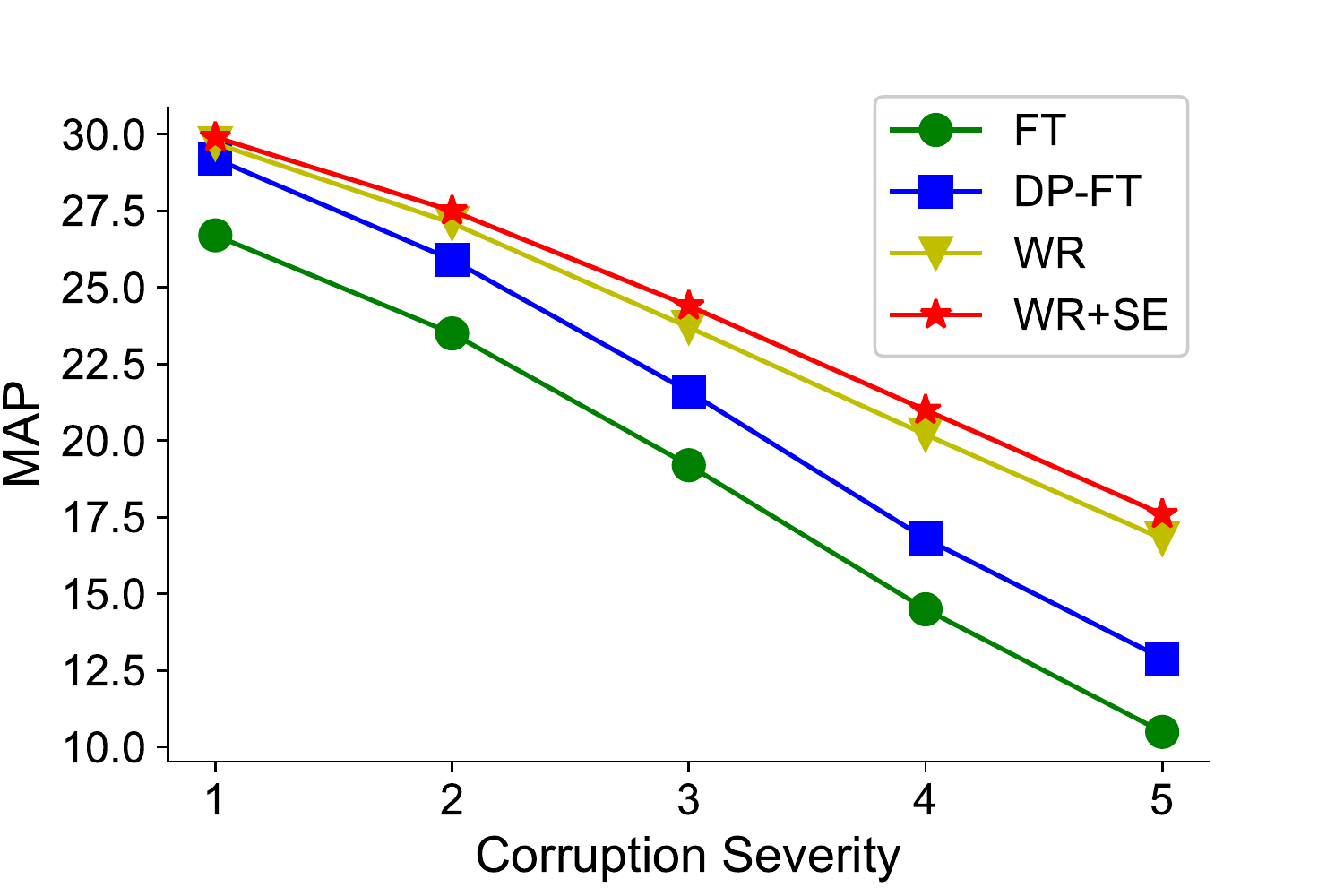}
    \caption{Performance on each corruption severity.}
    \label{fig:severity}
\end{figure}

%% file: figures/ex_pascal.tex
\begin{figure*}
    \centering
    \includegraphics[width=0.85\linewidth]{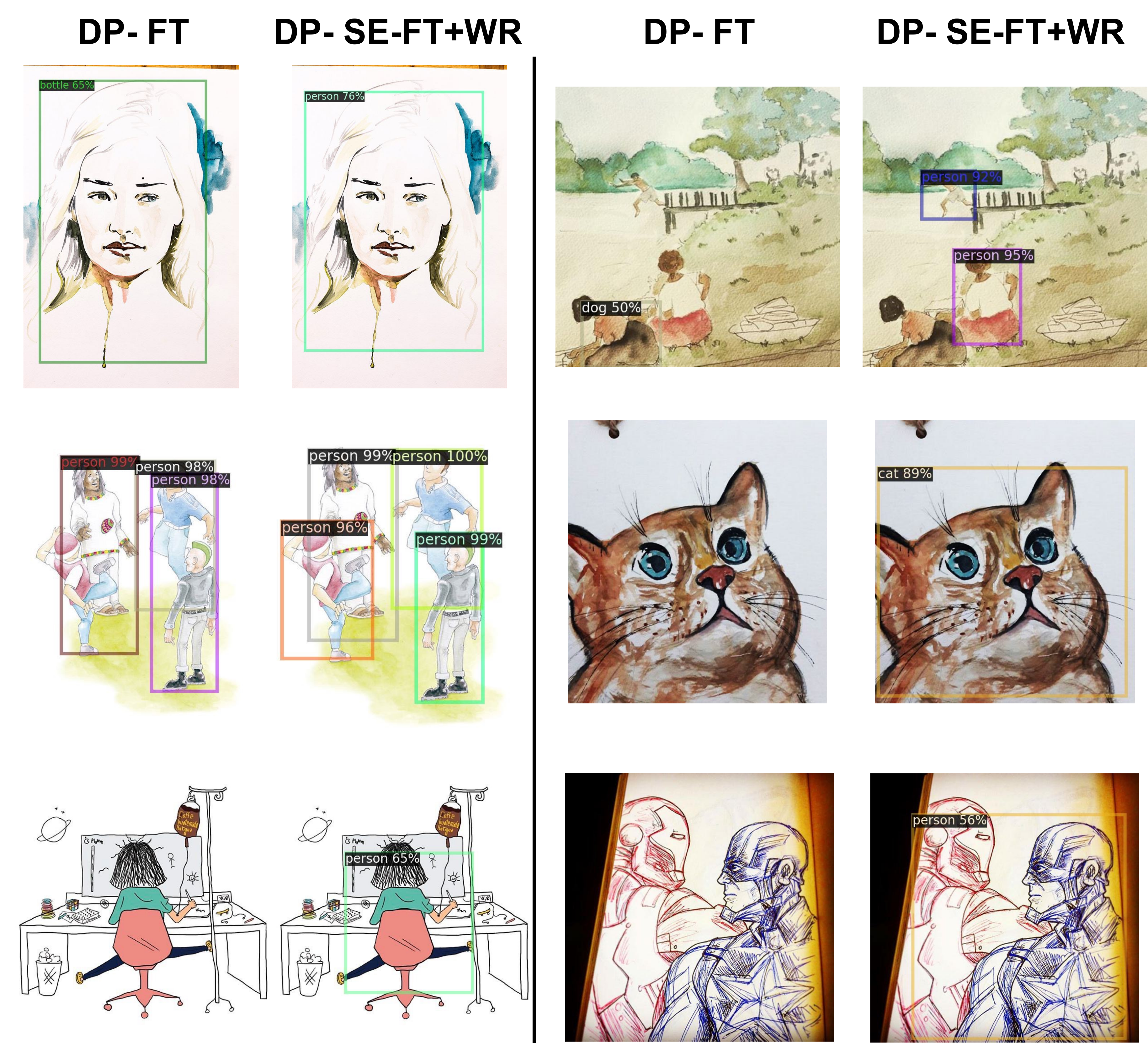}
    \caption{Detection results on Pascal.}
    \label{fig:ex_pascal}
\end{figure*}

%% file: figures/ex_cityscape.tex
\begin{figure*}[!ht]
    \centering
    \includegraphics[width=\linewidth]{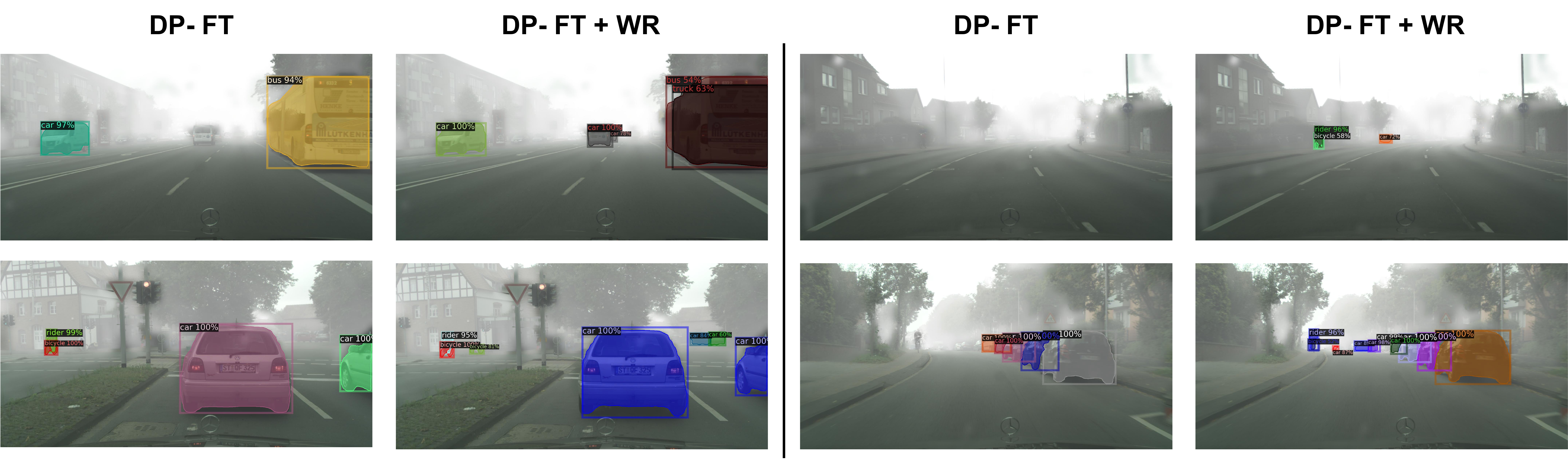}
    \caption{Detection results on Cityscapes.}
    \label{fig:ex_cityscape}
\end{figure*}

%% file: main_arxiv.bbl
\begin{thebibliography}{10}\itemsep=-1pt

\bibitem{angarano2022back}
Simone Angarano, Mauro Martini, Francesco Salvetti, Vittorio Mazzia, and
  Marcello Chiaberge.
\newblock Back-to-bones: Rediscovering the role of backbones in domain
  generalization.
\newblock {\em arXiv preprint arXiv:2209.01121}, 2022.

\bibitem{cha2021swad}
Junbum Cha, Sanghyuk Chun, Kyungjae Lee, Han-Cheol Cho, Seunghyun Park, Yunsung
  Lee, and Sungrae Park.
\newblock Swad: Domain generalization by seeking flat minima.
\newblock {\em Advances in Neural Information Processing Systems},
  34:22405--22418, 2021.

\bibitem{cha2022domain}
Junbum Cha, Kyungjae Lee, Sungrae Park, and Sanghyuk Chun.
\newblock Domain generalization by mutual-information regularization with
  pre-trained models.
\newblock In {\em Computer Vision--ECCV 2022: 17th European Conference, Tel
  Aviv, Israel, October 23--27, 2022, Proceedings, Part XXIII}, pages 440--457.
  Springer, 2022.

\bibitem{chen2021contrastive}
Wuyang Chen, Zhiding Yu, Shalini De~Mello, Sifei Liu, Jose~M Alvarez, Zhangyang
  Wang, and Anima Anandkumar.
\newblock Contrastive syn-to-real generalization.
\newblock {\em arXiv preprint arXiv:2104.02290}, 2021.

\bibitem{chen2021robust}
Xiangning Chen, Cihang Xie, Mingxing Tan, Li Zhang, Cho-Jui Hsieh, and Boqing
  Gong.
\newblock Robust and accurate object detection via adversarial learning.
\newblock In {\em Proceedings of the IEEE/CVF conference on computer vision and
  pattern recognition}, pages 16622--16631, 2021.

\bibitem{chu2016best}
Brian Chu, Vashisht Madhavan, Oscar Beijbom, Judy Hoffman, and Trevor Darrell.
\newblock Best practices for fine-tuning visual classifiers to new domains.
\newblock In {\em Computer Vision--ECCV 2016 Workshops: Amsterdam, The
  Netherlands, October 8-10 and 15-16, 2016, Proceedings, Part III 14}, pages
  435--442. Springer, 2016.

\bibitem{cordts2016cityscapes}
Marius Cordts, Mohamed Omran, Sebastian Ramos, Timo Rehfeld, Markus Enzweiler,
  Rodrigo Benenson, Uwe Franke, Stefan Roth, and Bernt Schiele.
\newblock The cityscapes dataset for semantic urban scene understanding.
\newblock In {\em CVPR}, 2016.

\bibitem{deng2009imagenet}
Jia Deng, Wei Dong, Richard Socher, Li-Jia Li, Kai Li, and Li Fei-Fei.
\newblock Imagenet: A large-scale hierarchical image database.
\newblock In {\em 2009 IEEE conference on computer vision and pattern
  recognition}, pages 248--255. Ieee, 2009.

\bibitem{everingham2010pascal}
Mark Everingham, Luc Van~Gool, Christopher~KI Williams, John Winn, and Andrew
  Zisserman.
\newblock The pascal visual object classes (voc) challenge.
\newblock {\em IJCV}, 88(2):303--338, 2010.

\bibitem{fan2022normalization}
Qi Fan, Mattia Segu, Yu-Wing Tai, Fisher Yu, Chi-Keung Tang, Bernt Schiele, and
  Dengxin Dai.
\newblock Normalization perturbation: A simple domain generalization method for
  real-world domain shifts.
\newblock {\em arXiv preprint arXiv:2211.04393}, 2022.

\bibitem{ghiasi2019fpn}
Golnaz Ghiasi, Tsung-Yi Lin, and Quoc~V Le.
\newblock Nas-fpn: Learning scalable feature pyramid architecture for object
  detection.
\newblock In {\em Proceedings of the IEEE/CVF conference on computer vision and
  pattern recognition}, pages 7036--7045, 2019.

\bibitem{girshick2014rich}
Ross Girshick, Jeff Donahue, Trevor Darrell, and Jitendra Malik.
\newblock Rich feature hierarchies for accurate object detection and semantic
  segmentation.
\newblock In {\em Proceedings of the IEEE conference on computer vision and
  pattern recognition}, pages 580--587, 2014.

\bibitem{gulrajani2020search}
Ishaan Gulrajani and David Lopez-Paz.
\newblock In search of lost domain generalization.
\newblock {\em arXiv preprint arXiv:2007.01434}, 2020.

\bibitem{gupta2019lvis}
Agrim Gupta, Piotr Dollar, and Ross Girshick.
\newblock Lvis: A dataset for large vocabulary instance segmentation.
\newblock In {\em Proceedings of the IEEE/CVF conference on computer vision and
  pattern recognition}, pages 5356--5364, 2019.

\bibitem{he2016deep}
Kaiming He, Xiangyu Zhang, Shaoqing Ren, and Jian Sun.
\newblock Deep residual learning for image recognition.
\newblock In {\em CVPR}, 2016.

\bibitem{hendrycks2019augmix}
Dan Hendrycks, Norman Mu, Ekin~D Cubuk, Barret Zoph, Justin Gilmer, and Balaji
  Lakshminarayanan.
\newblock Augmix: A simple data processing method to improve robustness and
  uncertainty.
\newblock {\em arXiv preprint arXiv:1912.02781}, 2019.

\bibitem{hu2018squeeze}
Jie Hu, Li Shen, and Gang Sun.
\newblock Squeeze-and-excitation networks.
\newblock In {\em Proceedings of the IEEE conference on computer vision and
  pattern recognition}, pages 7132--7141, 2018.

\bibitem{inoue2018cross}
Naoto Inoue, Ryosuke Furuta, Toshihiko Yamasaki, and Kiyoharu Aizawa.
\newblock Cross-domain weakly-supervised object detection through progressive
  domain adaptation.
\newblock In {\em CVPR}, 2018.

\bibitem{kim2022broad}
Donghyun Kim, Kaihong Wang, Stan Sclaroff, and Kate Saenko.
\newblock A broad study of pre-training for domain generalization and
  adaptation.
\newblock In {\em Computer Vision--ECCV 2022: 17th European Conference, Tel
  Aviv, Israel, October 23--27, 2022, Proceedings, Part XXXIII}, pages
  621--638. Springer, 2022.

\bibitem{kirkpatrick2016overcoming}
James Kirkpatrick, Razvan Pascanu, Neil Rabinowitz, Joel Veness, Guillaume
  Desjardins, Andrei~A Rusu, Kieran Milan, John Quan, Tiago Ramalho, Agnieszka
  Grabska-Barwinska, et~al.
\newblock Overcoming catastrophic forgetting in neural networks.
\newblock {\em arXiv preprint arXiv:1612.00796}, 2016.

\bibitem{kumar2022fine}
Ananya Kumar, Aditi Raghunathan, Robbie Jones, Tengyu Ma, and Percy Liang.
\newblock Fine-tuning can distort pretrained features and underperform
  out-of-distribution.
\newblock {\em arXiv preprint arXiv:2202.10054}, 2022.

\bibitem{lee2022surgical}
Yoonho Lee, Annie~S Chen, Fahim Tajwar, Ananya Kumar, Huaxiu Yao, Percy Liang,
  and Chelsea Finn.
\newblock Surgical fine-tuning improves adaptation to distribution shifts.
\newblock {\em arXiv preprint arXiv:2210.11466}, 2022.

\bibitem{lin2021domain}
Chuang Lin, Zehuan Yuan, Sicheng Zhao, Peize Sun, Changhu Wang, and Jianfei
  Cai.
\newblock Domain-invariant disentangled network for generalizable object
  detection.
\newblock In {\em Proceedings of the IEEE/CVF International Conference on
  Computer Vision}, pages 8771--8780, 2021.

\bibitem{lin2017feature}
Tsung-Yi Lin, Piotr Doll{\'a}r, Ross Girshick, Kaiming He, Bharath Hariharan,
  and Serge Belongie.
\newblock Feature pyramid networks for object detection.
\newblock In {\em CVPR}, pages 2117--2125, 2017.

\bibitem{coco}
Tsung-Yi Lin, Michael Maire, Serge Belongie, James Hays, Pietro Perona, Deva
  Ramanan, Piotr Doll{\'a}r, and C~Lawrence Zitnick.
\newblock Microsoft coco: Common objects in context.
\newblock In {\em ECCV}, 2014.

\bibitem{liu2021swin}
Ze Liu, Yutong Lin, Yue Cao, Han Hu, Yixuan Wei, Zheng Zhang, Stephen Lin, and
  Baining Guo.
\newblock Swin transformer: Hierarchical vision transformer using shifted
  windows.
\newblock In {\em Proceedings of the IEEE/CVF international conference on
  computer vision}, pages 10012--10022, 2021.

\bibitem{liu2022convnet}
Zhuang Liu, Hanzi Mao, Chao-Yuan Wu, Christoph Feichtenhofer, Trevor Darrell,
  and Saining Xie.
\newblock A convnet for the 2020s.
\newblock In {\em Proceedings of the IEEE/CVF Conference on Computer Vision and
  Pattern Recognition}, pages 11976--11986, 2022.

\bibitem{long2015fully}
Jonathan Long, Evan Shelhamer, and Trevor Darrell.
\newblock Fully convolutional networks for semantic segmentation.
\newblock In {\em Proceedings of the IEEE conference on computer vision and
  pattern recognition}, pages 3431--3440, 2015.

\bibitem{mahajan2018exploring}
Dhruv Mahajan, Ross Girshick, Vignesh Ramanathan, Kaiming He, Manohar Paluri,
  Yixuan Li, Ashwin Bharambe, and Laurens Van Der~Maaten.
\newblock Exploring the limits of weakly supervised pretraining.
\newblock In {\em Proceedings of the European conference on computer vision
  (ECCV)}, pages 181--196, 2018.

\bibitem{michaelis2019benchmarking}
Claudio Michaelis, Benjamin Mitzkus, Robert Geirhos, Evgenia Rusak, Oliver
  Bringmann, Alexander~S Ecker, Matthias Bethge, and Wieland Brendel.
\newblock Benchmarking robustness in object detection: Autonomous driving when
  winter is coming.
\newblock {\em arXiv preprint arXiv:1907.07484}, 2019.

\bibitem{mirzadeh2022wide}
Seyed~Iman Mirzadeh, Arslan Chaudhry, Dong Yin, Huiyi Hu, Razvan Pascanu, Dilan
  Gorur, and Mehrdad Farajtabar.
\newblock Wide neural networks forget less catastrophically.
\newblock In {\em International Conference on Machine Learning}, pages
  15699--15717. PMLR, 2022.

\bibitem{rame2022diverse}
Alexandre Rame, Matthieu Kirchmeyer, Thibaud Rahier, Alain Rakotomamonjy,
  Patrick Gallinari, and Matthieu Cord.
\newblock Diverse weight averaging for out-of-distribution generalization.
\newblock {\em arXiv preprint arXiv:2205.09739}, 2022.

\bibitem{faster}
Shaoqing Ren, Kaiming He, Ross Girshick, and Jian Sun.
\newblock Faster r-cnn: Towards real-time object detection with region proposal
  networks.
\newblock In {\em NeurIPS}, 2015.

\bibitem{sakaridis2018semantic}
Christos Sakaridis, Dengxin Dai, and Luc Van~Gool.
\newblock Semantic foggy scene understanding with synthetic data.
\newblock {\em IJCV}, 2018.

\bibitem{sandler2018mobilenetv2}
Mark Sandler, Andrew Howard, Menglong Zhu, Andrey Zhmoginov, and Liang-Chieh
  Chen.
\newblock Mobilenetv2: Inverted residuals and linear bottlenecks.
\newblock In {\em Proceedings of the IEEE conference on computer vision and
  pattern recognition}, pages 4510--4520, 2018.

\bibitem{sun2017revisiting}
Chen Sun, Abhinav Shrivastava, Saurabh Singh, and Abhinav Gupta.
\newblock Revisiting unreasonable effectiveness of data in deep learning era.
\newblock In {\em Proceedings of the IEEE international conference on computer
  vision}, pages 843--852, 2017.

\bibitem{tan2019efficientnet}
Mingxing Tan and Quoc Le.
\newblock Efficientnet: Rethinking model scaling for convolutional neural
  networks.
\newblock In {\em International conference on machine learning}, pages
  6105--6114. PMLR, 2019.

\bibitem{vasconcelos2022proper}
Cristina Vasconcelos, Vighnesh Birodkar, and Vincent Dumoulin.
\newblock Proper reuse of image classification features improves object
  detection.
\newblock In {\em Proceedings of the IEEE/CVF Conference on Computer Vision and
  Pattern Recognition}, pages 13628--13637, 2022.

\bibitem{wang2019towards}
Xudong Wang, Zhaowei Cai, Dashan Gao, and Nuno Vasconcelos.
\newblock Towards universal object detection by domain attention.
\newblock In {\em Proceedings of the IEEE/CVF conference on computer vision and
  pattern recognition}, pages 7289--7298, 2019.

\bibitem{wang2021robust}
Xin Wang, Thomas~E Huang, Benlin Liu, Fisher Yu, Xiaolong Wang, Joseph~E
  Gonzalez, and Trevor Darrell.
\newblock Robust object detection via instance-level temporal cycle confusion.
\newblock {\em International Conference on Computer Vision (ICCV)}, 2021.

\bibitem{rw2019timm}
Ross Wightman.
\newblock Pytorch image models.
\newblock \url{https://github.com/rwightman/pytorch-image-models}, 2019.

\bibitem{wortsman2022model}
Mitchell Wortsman, Gabriel Ilharco, Samir~Ya Gadre, Rebecca Roelofs, Raphael
  Gontijo-Lopes, Ari~S Morcos, Hongseok Namkoong, Ali Farhadi, Yair Carmon,
  Simon Kornblith, et~al.
\newblock Model soups: averaging weights of multiple fine-tuned models improves
  accuracy without increasing inference time.
\newblock In {\em International Conference on Machine Learning}, pages
  23965--23998. PMLR, 2022.

\bibitem{wortsman2022robust}
Mitchell Wortsman, Gabriel Ilharco, Jong~Wook Kim, Mike Li, Simon Kornblith,
  Rebecca Roelofs, Raphael~Gontijo Lopes, Hannaneh Hajishirzi, Ali Farhadi,
  Hongseok Namkoong, et~al.
\newblock Robust fine-tuning of zero-shot models.
\newblock In {\em Proceedings of the IEEE/CVF Conference on Computer Vision and
  Pattern Recognition}, pages 7959--7971, 2022.

\bibitem{wu2018group}
Yuxin Wu and Kaiming He.
\newblock Group normalization.
\newblock In {\em Proceedings of the European conference on computer vision
  (ECCV)}, pages 3--19, 2018.

\bibitem{wu2019detectron2}
Yuxin Wu, Alexander Kirillov, Francisco Massa, Wan-Yen Lo, and Ross Girshick.
\newblock Detectron2.
\newblock \url{https://github.com/facebookresearch/detectron2}, 2019.

\bibitem{xie2020self}
Qizhe Xie, Minh-Thang Luong, Eduard Hovy, and Quoc~V Le.
\newblock Self-training with noisy student improves imagenet classification.
\newblock In {\em Proceedings of the IEEE/CVF conference on computer vision and
  pattern recognition}, pages 10687--10698, 2020.

\bibitem{xu2021fourier}
Qinwei Xu, Ruipeng Zhang, Ya Zhang, Yanfeng Wang, and Qi Tian.
\newblock A fourier-based framework for domain generalization.
\newblock In {\em Proceedings of the IEEE/CVF Conference on Computer Vision and
  Pattern Recognition}, pages 14383--14392, 2021.

\bibitem{xuhong2018explicit}
LI Xuhong, Yves Grandvalet, and Franck Davoine.
\newblock Explicit inductive bias for transfer learning with convolutional
  networks.
\newblock In {\em International Conference on Machine Learning}, pages
  2825--2834. PMLR, 2018.

\bibitem{yosinski2014transferable}
Jason Yosinski, Jeff Clune, Yoshua Bengio, and Hod Lipson.
\newblock How transferable are features in deep neural networks?
\newblock {\em Advances in neural information processing systems}, 27, 2014.

\bibitem{yu2020bdd100k}
Fisher Yu, Haofeng Chen, Xin Wang, Wenqi Xian, Yingying Chen, Fangchen Liu,
  Vashisht Madhavan, and Trevor Darrell.
\newblock Bdd100k: A diverse driving dataset for heterogeneous multitask
  learning.
\newblock In {\em Proceedings of the IEEE/CVF conference on computer vision and
  pattern recognition}, pages 2636--2645, 2020.

\bibitem{zhong2022adversarial}
Zhun Zhong, Yuyang Zhao, Gim~Hee Lee, and Nicu Sebe.
\newblock Adversarial style augmentation for domain generalized urban-scene
  segmentation.
\newblock {\em arXiv preprint arXiv:2207.04892}, 2022.

\bibitem{zhou2021domain}
Kaiyang Zhou, Yongxin Yang, Yu Qiao, and Tao Xiang.
\newblock Domain generalization with mixstyle.
\newblock {\em arXiv preprint arXiv:2104.02008}, 2021.

\end{thebibliography}
